\documentclass[10pt,twocolumn,letterpaper]{article}

\usepackage[pagenumbers]{cvpr} % To force page numbers, e.g. for an arXiv version

\usepackage{graphicx}
\usepackage{amsmath,bm}
\usepackage{amssymb}
\usepackage{booktabs}
\usepackage{algorithmic}
\usepackage[linesnumbered,ruled,vlined]{algorithm2e}
\SetKwInput{KwInput}{Input}                % Set the Input
\SetKwInput{KwOutput}{Output}              % set the Output
\usepackage{eucal}

\DeclareMathOperator*{\E}{\mathbb{E}}

\usepackage{epigraph}
\usepackage[table,xcdraw]{xcolor}
\usepackage{wrapfig,lipsum,booktabs}
\usepackage{enumitem}

\usepackage[pagebackref=true,breaklinks=true,letterpaper=true,colorlinks,urlcolor=black,bookmarks=false]{hyperref}
\hypersetup{citecolor=[RGB]{119,185,0}}

\usepackage[capitalize]{cleveref}
\crefname{section}{Sec.}{Secs.}
\Crefname{section}{Section}{Sections}
\Crefname{table}{Table}{Tables}
\crefname{table}{Tab.}{Tabs.}

\def\cvprPaperID{943} % *** Enter the CVPR Paper ID here
\def\confName{CVPR}
\def\confYear{2022}

\begin{document}

%%%%%%%%% TITLE - PLEASE UPDATE
% \title{Reducing Labeling Cost in Deep Object Detection}
\title{Not All Labels Are Created Equal: A Story of Labeling Cost for Object Detection}
%\title{Not All Labels Are Created Equal: A Study of Labeling Cost for Object Detection}
\title{Not All Labels Are Equal:\\
Rationalizing The Labeling Costs for Training Object Detection}

\author{Ismail Elezi\textsuperscript{1}\thanks{Work  performed while interning at NVIDIA.} \quad  Zhiding Yu\textsuperscript{2} \quad  Anima Anandkumar\textsuperscript{2,3} \quad  Laura Leal-Taix\'{e}\textsuperscript{1} \quad Jose M. Alvarez\textsuperscript{2}\\
	\textsuperscript{1}TUM \hspace{1cm}
	\textsuperscript{2}NVIDIA \hspace{1cm} 
	\textsuperscript{3}CALTECH  \\
}
\maketitle

%%%%%%%%% ABSTRACT
\begin{abstract}
Deep neural networks have reached high accuracy on object detection but their success hinges on large amounts of labeled data. 
To reduce the labels dependency, various active learning strategies have been proposed, typically based on the confidence of the detector. However, these methods are biased towards high-performing classes and can lead to acquired datasets that are not good representatives of the testing set data.
In this work, we propose a unified framework for active learning, that considers both the uncertainty and the robustness of the detector, ensuring that the network performs well in all classes. Furthermore, our method leverages auto-labeling to suppress a potential distribution drift while boosting the performance of the model. Experiments on PASCAL VOC07+12 and MS-COCO show that our method consistently outperforms a wide range of active learning methods, yielding up to a $7.7\%$  improvement in mAP, or up to $82\%$ reduction in labeling cost. Code will be released upon acceptance of the paper.
\end{abstract}

%our method is able to pseudo-label confident predictions, suppressing a potential distribution drift while boosting the performance of the model.

%\epigraph{“All labels are equal, but some are more equal than others.”}{\textit{Unknown \cite{label_farm}}}

%%%%%%%%% BODY TEXT
\vspace{-2mm}
\section{Introduction}
The performance of deep object detection networks \cite{DBLP:conf/nips/RenHGS15, liu2016ssd} depends heavily on the size of the labeled dataset. Adding more labeled data helps, yet adding more data costs. 
Therefore, it is imperative to adopt active learning (AL) strategies to select the most informative samples in the dataset for labeling, and self and semi-supervised learning (SSL) approaches to leverage unlabeled data whenever possible.

\iffalse
\begin{figure}[t!]
\centering
\includegraphics[scale=0.12]{images/teaser_new_new_new.png}
    \caption{An object from a low-performing class (Pottedplant). a) Because of its low-entropy, uncertainty-based AL methods do not label the image. b) Because of its high confidence, pseudo-label SSL methods wrongly pseudo-label the image and thus harm the training. c) Because of its high-inconsistency, consistency-based SSL methods cannot learn from it.  d) Our method selects the image for labeling and prevents it from getting pseudo-labeled. The blue color represents the original image, the orange color represents its augmented version.}
    \vspace{0.1cm}
\label{fig:teaser}
\end{figure}
\fi

\begin{figure}[t!]
\centering
\includegraphics[width=1\linewidth]{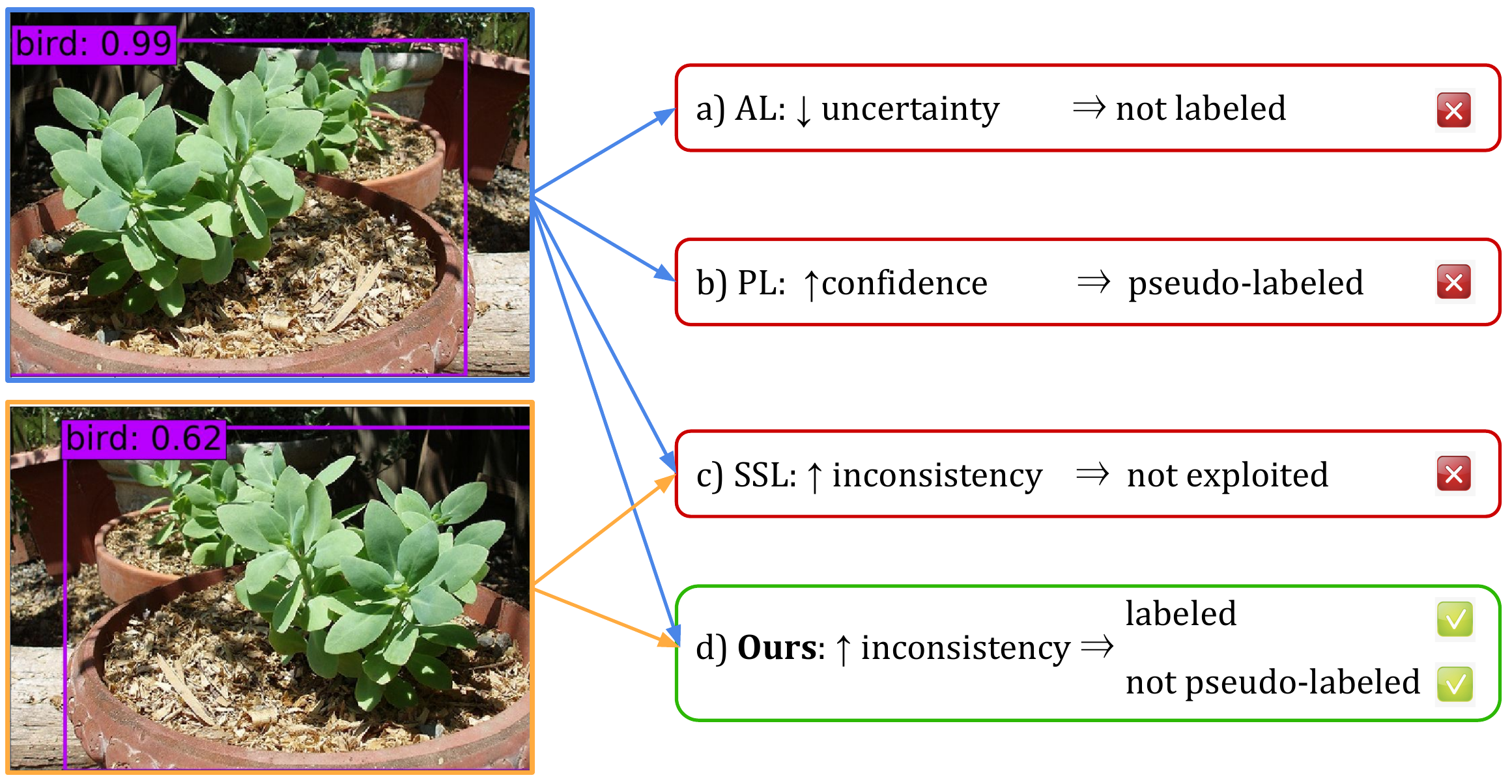}
    \caption{An object from a low-performing class (Pottedplant), original image in blue, augmented version in orange. a) Because of its low-entropy, uncertainty-based AL methods do not label the image. b) Because of its high confidence, pseudo-label SSL methods wrongly pseudo-label the image and thus harm the training. c) Because of its high-inconsistency, consistency-based SSL methods cannot learn from it.  d) Our method selects the image for labeling and prevents it from getting pseudo-labeled.}
    \vspace{-0.4cm}
\label{fig:teaser}
\end{figure}

Consistency-based Semi-Supervised Learning (SSL) methods for object detection~\cite{DBLP:conf/nips/JeongLKK19} train a network to minimize the inconsistency between its predictions. However, as shown in Fig. \ref{fig:teaser}, some images still give inconsistent predictions, and thus the network does not learn from them. Auto-labeling uses self-learning to label high confident predictions, \ie pseudo-label (PL), but, since networks are miscalibrated, they can generate wrong labels, potentially harming the training. Moreover, by targetting high-confident predictions, they ignore objects of low-performing classes.

When samples cannot be pseudo-labeled, an alternative is to obtain the ground truth via manual labeling. Active learning (AL) for object detection~\cite{DBLP:conf/cvpr/YooK19, DBLP:journals/corr/abs-2103-16130} is a common approach to select the most promising samples for labeling to reduce labeling costs. The selection is based on an acquisition function to assess the informativeness of an image, typically computed based on the network's uncertainty.
However, the acquisition function is only meaningful if the network is already well-trained for the task, which is not always the case, especially in the early AL cycles. Even if the network performs well in most classes, significant intra-class variance can lead to a low accuracy on a particular class, see Fig. \ref{fig:teaser}a. 
In those cases, using the network predictions to compute the acquisition function can lead to worse performance than random sampling. Furthermore, we change the dataset distribution at every AL cycle by selecting only the most uncertain (hard) samples until they no longer resemble the test distribution.

In this work, we advocate for a holistic view of the labeling problem, that is, a unified strategy to choose which samples to manually label and which samples can be automatically labeled.
We start from an \textit{uncertainty}-based AL framework and generalize its acquisition function by introducing the concept of \textit{robustness}, which is commonly not present on AL. If a network's predictions of an image and its random augmentations, i.e., horizontal flipping, are not consistent, the image needs to be manually labeled. This simple yet effective change allows us to select informative samples for both low and high-performing classes.
This is unlike classic SSL settings, where samples with inconsistent predictions would neither be labeled nor pseudo-labeled, hence, the information contained in them would not be used.

We are still left with the potential dataset distribution drift, for which we propose to use auto-labeling in order to not increase the labeling costs.
For every active learning cycle, we use the previously trained network to mine easy samples, i.e., samples where the network is confident about its prediction, and use the network's own prediction as labels. 
Note, that easy samples are typically not used in AL cycles. Only by holistically thinking about which samples to manually label and which to auto-label can we take full advantage of the entire dataset. In summary, our \textbf{contributions} are the following:
\begin{itemize}[leftmargin=*]
\item We propose a novel class-agnostic active learning score based on the \textit{robustness} of the network, using a novel \textit{inconsistency} score. 
\item We use auto-labeling to leverage the less informative samples, expanding the labeled dataset for free.
\item We demonstrate the benefits of our method in two publicly available datasets: PASCAL VOC07+12 and MS-COCO. Compared to state-of-the-art methods \cite{DBLP:conf/iclr/SenerS18, DBLP:conf/cvpr/YooK19, DBLP:conf/eccv/Agarwal0AA20, DBLP:conf/cvpr/YuanWFLXJY21, DBLP:journals/corr/abs-2103-16130}, our approach yields up to a $7.7\%$ and $7\%$ relative mAP improvement for PASCAL-VOC and MS-COCO, respectively. Importantly, we can achieve the same performance as the baseline but reduce up to $82\%$ of the labeling costs.
\end{itemize}

\iffalse
\begin{figure*}[t!]
\centering
\includegraphics[scale=0.4]{images/pipeline_nice.png}
\caption{Overview of our method. We first train the network in a semi-supervised manner. During active learning, for each image, we use the network to compute the acquisition function and based on it decide if we actively label the image, if we pseudo-label it, or if we only use it as part of the unlabeled data for the next training cycle.}
\label{fig:pipeline}
\vspace{0.2cm}
\end{figure*}
\fi

\begin{figure*}[t!]
\centering
\vspace{-1mm}
\begin{minipage}{0.735\linewidth}
% \centering
\vspace{-4mm}
\includegraphics[width=0.99\linewidth]{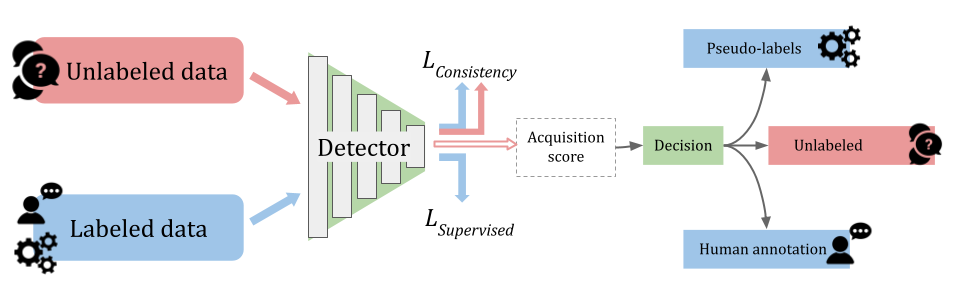} 
\end{minipage}
\begin{minipage}{0.015\linewidth}
\end{minipage}
\begin{minipage}{0.255\linewidth}
\caption{Overview of our method. We first train the network in a semi-supervised manner. During active learning, for each image, we use the network to compute the acquisition function and based on it decide if we actively label the image, if we pseudo-label it, or if we only use it as part of the unlabeled data for the next training cycle.}
\label{fig:pipeline} 
\end{minipage}
\vspace{-5mm}
\end{figure*}
\section{Related Work}
\textbf{Deep Active Learning (AL) for Object Detection.} 
The traditional way of doing AL is by training an ensemble of neural networks \cite{DBLP:conf/cvpr/BeluchGNK18} and then selecting the samples with the highest score defined by some acquisition function, i.e., entropy \cite{DBLP:journals/sigmobile/Shannon01}, or BALD \cite{DBLP:journals/corr/abs-1112-5745}. Concurrent works \cite{DBLP:conf/icml/GalIG17,DBLP:conf/nips/KirschAG19} explore a similar direction by approximating the uncertainty via Monte-Carlo dropout \cite{DBLP:conf/icml/GalG16}. These approaches are compared \cite{DBLP:conf/cvpr/BeluchGNK18}, with the authors concluding that the ensemble approach reaches higher results at the cost of more computational power. Another Bayesian approach \cite{DBLP:conf/icml/TranDRC19} trains a variational autoencoder (VAE) \cite{DBLP:journals/corr/KingmaW13} on both real and augmented samples, and then chooses to label the samples with the highest reconstruction error. A different approach is the core-set \cite{DBLP:conf/iclr/SenerS18} that chooses to label a set of points such that a model trained over the selected subset is competitive for the remaining data points. In the work of \cite{DBLP:conf/eccv/GaoZYADP20} the authors tailor the method of \cite{DBLP:conf/nips/BerthelotCGPOR19} for AL. In general,  these methods are designed for classification-based AL and can not be directly applied to object detection.%it is not trivial to use them for object detection.

Recently, several methods have been adapted specifically for the task of object detection \cite{DBLP:conf/iccv/AghdamGLW19, DBLP:conf/accv/KaoLS018, DBLP:conf/bmvc/RoyUN18, DBLP:conf/bmvc/DesaiLGNB19, DBLP:conf/iclr/SenerS18, DBLP:conf/eccv/Agarwal0AA20}, some of which are based on the core-set approaches where the diversity of the training examples is taken into account. 
However, the state-of-the-art approaches are based on the uncertainty \cite{haussmann2020scalable, DBLP:conf/cvpr/YooK19, DBLP:conf/cvpr/YuanWFLXJY21, DBLP:journals/corr/abs-2103-16130}. 
The work of \cite{haussmann2020scalable} consists of an ensemble of object detectors that provide bounding boxes and probabilities for each class of interest. Then, a scoring function is used to obtain a single value representing the informativeness of each unlabeled image. Similar to that is the work of \cite{DBLP:conf/cvpr/YuanWFLXJY21} where the authors compute the instance-based uncertainty. Another work ~\cite{DBLP:conf/cvpr/YooK19} gives an elegant solution, reaching promising results compared with other single-model methods. The authors train a network in the task of detection while learning to predict the final loss. In the sample acquisition stage, samples with the highest prediction loss are considered the most interesting ones and are chosen to be labeled. In the state-of-the-art approach \cite{DBLP:journals/corr/abs-2103-16130}, authors define the aleatoric and epistemic uncertainty, in both class and bounding box level, and use the combined score to determine the images that need labeling. Our work is related but different from the above-mentioned works. Similarly, we consider the uncertainty of the detector as part of the solution. Unlike them, we find that the robustness of the detector is even more reliable as an acquisition function, especially for the low-performing classes. We then unify these two scores to reach high performance in the majority of classes.

\textbf{Deep Semi-Supervised Learning (SSL) for Object Detection} is a deep learning approach that combines a small amount of labeled data with a large amount of unlabeled data during neural network training. Unlike in AL, where the unlabeled data is used only during the acquiring stage, in SSL, the unlabeled data are used during the training. Several methods have shown excellent results~\cite{pseudo-labelling,DBLP:conf/nips/TarvainenV17,DBLP:conf/iclr/LaineA17,DBLP:journals/pami/MiyatoMKI19} by casting the problem of semi-supervised learning as a regularization problem, in effect adding a new loss for the unlabeled samples. %
Follow-up works significantly improved the performance of SSL in object classification \cite{DBLP:conf/nips/OliverORCG18, DBLP:conf/nips/BerthelotCGPOR19, DBLP:conf/iclr/BerthelotCCKSZR20, DBLP:conf/nips/SohnBCZZRCKL20, DBLP:journals/corr/abs-2110-08263, DBLP:journals/corr/abs-2106-04732}. 

While going from semi-supervised image classification to semi-supervised object detection is challenging, some promising directions are given in recent works \cite{DBLP:conf/nips/JeongLKK19, DBLP:conf/cvpr/JeongVHKK21, DBLP:conf/cvpr/TangCLZ21} where the authors develop loss functions that minimize the inconsistency between images and their augmented version. The methods significantly improved the mAP score in Pascal VOC dataset \cite{DBLP:journals/ijcv/EveringhamGWWZ10}. Our work is inspired by \cite{DBLP:conf/nips/JeongLKK19} but we instead develop an acquisition function that computes the inconsistency between an image and an augmented version of it, and show that such a score is more reliable than the uncertainty, especially for low-performing classes. Furthermore, we add a pseudo-labeling module that labels the easy images for free. Together with our developed AL method, it ensures that our acquired dataset is a good representative of the original dataset, in turn, improving the results.

\section{Method}
Let $D$ be a dataset divided into a labeled set $L$ and a pool of unlabeled data $U$. 
We describe our acquisition function for AL in Sec. \ref{sec:mining}. This consists of mining a subset of samples from the pool of unlabeled data $U$ and transferring them to the labeled set $L$, incurring a labeling cost. 
However, arbitrarily augmenting the set $L$ with only hard samples creates a distribution drift in our training data. 
Hence, we propose to include in training the easy samples, i.e., objects for which the network's confidence is high, by using pseudo-labeling (Sec. \ref{pseudo-labeling}). 
Finally, we describe the training procedure in Sec. \ref{training}.
We show a high-level description of the method in Fig.~\ref{fig:pipeline}.

\textbf{Notation.}\label{notation}
Let $\bm{\Delta}$ be the object predictions for an image, and let $\bm{\Delta}_i$ be its i-th object prediction. $\bm{\Delta}_i$ consists of the bounding box $\bm{b}_i$ and $\bm{c}_i$ represents the probability distribution after the softmax layer of the neural network. We denote the $p$-th category of the distribution with $\bm{c}_i^p$. The bounding box $\bm{b}_i$ consists of the \textit{displacement of the center} and \textit{scale coefficients}, represented by the tuple $[\bm{\delta x}, \bm{\delta y}, \bm{w}, \bm{h}]$. Given a weakly augmented version of the original image, e.g., by doing a horizontal flip, we define $\bm{\hat{\Delta}}$ to be the set of its object predictions, and $\bm{\hat{\Delta}_i}$ consisting of the bounding box $\bm{\hat{b}}_i$ and $\bm{\hat{c}}_i$, its i-th prediction. 

\subsection{Inconsistency-based AL} 
\label{sec:mining}

Most AL methods use some measure of uncertainty, e.g., the entropy, to compute the acquisition function. A prediction that has a high entropy suggests that the object is highly dissimilar to the images the network is trained on. Thus, if labeled, it will provide different information to the ones we have. 
However, we empirically find that using only an \textit{uncertainty}-based acquisition function is not an ideal solution, especially for images coming from low-performing classes. As we show in the experiments, if the network's predictions for a class are incorrect, they are also unreliable to compute the acquisition function. 

\textbf{Inconsistency-based acquisition function.} To solve this issue, we propose a \textit{robustness}-based score for AL based on the \textit{consistency} between an image and its augmented version. If the predictions from an image and its augmented version are very similar, then we say that the network is robust for that image. 
On the other side, images where the network is inconsistent provide information that has not been captured in the training process and are prime candidates to be labeled. By focusing on robustness, the method is class-agnostic and performs well in most classes.

To measure the robustness of the network, we first define the inconsistency acquisition function $\mathcal{L}_{con_C}$. To do so, we feed the image and an augmented version of it to the detector. In our case, we use horizontal flip as augmentation.
Given the sets of predictions for the original and augmented image, we first need to match the predictions $\bm{\Delta}$ with $\bm{\hat{\Delta}}$. 
We do so by computing their intersection over union (IoU): 
\begin{equation}
\label{eq:matching}
\bm{\Delta'}_i = \text{argmax}_{\bm{b}_i \in \{\bm{b}\}} IoU(\bm{b}_i, \bm{\hat{b}}_i).
\end{equation}
For each matched pair, $\bm{\Delta}'_i$ and $\bm{\hat{\Delta}_i}$, we define their inconsistency as:
\begin{equation}
\label{eq:cls_loss}
\mathcal{L}_{con_C}(\bm{c'}_i, \bm{\hat{c}}_i) = \frac{1}{2} [KL(\bm{c'}_i, \bm{\hat{c}}_i) + KL(\bm{\hat{c}}_i, \bm{c'}_i)],
\end{equation}
\noindent where KL represents the Kullback-Leibler divergence. The higher the inconsistency, the more informative the sample is for training and therefore potentially worth labeling. 

\noindent \textbf{Aggregating object scores for image selection.} Given $\mathcal{L}_{con_C}$, the inconsistency for each object prediction in an image, we define the inconsistency of an image by aggregating the scores over $\bm{\Delta}$. Specifically, we first apply non-maximum suppression over its predictions, and then, define its inconsistency as:
\begin{equation}
    I(\bm{\Delta}) = max_i \{\mathcal{L}_{con_C}(\bm{c'}_i, \bm{\hat{c}}_i)\}.
\end{equation}

\noindent Similarly, we define the uncertainty of the image as:
\begin{equation}
    H(\bm{\Delta}) = max_i \{H(\bm{c}_i)\},
\end{equation}
\noindent where $H(\bm{c}_i)$ represents the entropy over distribution $\bm{c}_i$. The intuition behind using the \textit{maximum} score instead of some other score, such as the \textit{average}, is that labeling an image that has at least one \textit{difficult} object, independently of the number of \textit{easy} objects is beneficial because of the difficult object. %\cite{haussmann2020scalable}. 
Considering that the inconsistency and the uncertainty scores are on different scales, we unify them by multiplying them.
For $\bm{\Delta}$, we formulate our unified acquisition score as:

\begin{equation}
A(\bm{\Delta}) = H(\bm{\Delta}) \times I(\bm{\Delta}).
\label{eq:acquisition}
\end{equation}

Having scored each image in $U$, we then sort all the images based on their acquisition score and select to label the $N/T$ images with the highest score, where $N$ corresponds to the acquisition budget, and $T$ corresponds to the number of active learning steps. Note that we annotate every bounding box that belongs to a \textit{selected image} regardless of whether the box has a high score or not. We repeat this procedure for $T$ active learning cycles.

\subsection{Pseudo-labeling to prevent distribution drift}
\label{pseudo-labeling}
The active learning pipeline described above targets the most informative (hard) samples, ignoring the confident samples. We argue that the network should see some representative easy samples in order to ensure that no distribution drift happens. 
At the same time, we want to avoid labeling confident samples to not spend labeling resources.
Hence, we propose to use pseudo-labeling, where the network trained in the previous AL cycle provides pseudo-labels for the network that is being trained on this cycle. 
We pseudo-label an object if the network is confident, the confidence is above some threshold $\tau$:
\begin{equation}
	\hat{y}_i^p =
	\begin{cases}
		1, & \text{if } p=argmax(\bm{c}_i) \text{ and } \bm{c}_i^p \geq {\tau}\\
		0, & \text{otherwise.}
	\end{cases}
\label{eq:autolabel}
\end{equation}

We then use the one-hot pseudo-labels as ground truth for the training of the current network.
We note that in an image, the network might be confident for some predicted bounding box, and not confident for the others. As a toy example, given an image containing a cat and a dog, the network might be confident for the cat and pseudo-label it, but not confident for the dog. If we do not consider this, the standard loss functions will penalize the predictions of the network for the region of the image which contains the dog. However, because the region is unlabeled, the ground truth of that object will be set to background, in turn giving a high loss even if the network makes accurate prediction (as dog). In the next section, we describe how to fix this issue.

\subsection{Deep Object Detection Training} 
\label{training}
In this section, we describe the different losses used in our unified framework for training the deep object detector. First, we describe the multibox, consistency, and pseudo-labeling losses and finally the overall training loss.

\textbf{Multibox loss for labeled samples.} For the labeled images, the network is trained with the standard MultiBox loss for class predictions, and a smooth $\mathcal{L}\textit{1}$ loss for bounding box predictions. Given the network's class predictions $\bm{c}$ and an indicator $\bm{y_{ij}^p}=\{0,1\}$ for matching the $i$-th box to its corresponding $j$-th ground truth box of category $p$, the MultiBox loss is defined as \cite{liu2016ssd}:
\begin{small}
\begin{equation}
\label{loss:multibox}
\mathcal{L}_{conf}(\bm{c}, \bm{y}) = - \sum_{i \in Pos} \sum_{p=1}^{|\text{classes}|}  \bm{y}_{ij}^p log (\bm{c}_i^p)  - \sum_{i \in Neg} log (\bm{c}_i^0),
\end{equation}
\noindent where $Pos$ defines positive bounding boxes (containing objects), $Neg$ defines bounding boxes of class \textit{background}. % and $p$ defines the $p$-th category.
\end{small}

\textbf{Consistency loss for unlabeled samples.} Our approach leverages the inconsistency of the detector in the acquisition function. Intuitively, if the detector has high inconsistency in an image, it can not learn from it in a self-supervised manner, and the only way to learn from that image is to label it during the AL cycle. During training, we need to guide the detector to provide consistent predictions. To this end, we mirror the active learning procedure and feed an image and its augmented version to the detector using horizontal flips.  After matching the predictions, as described in Eq. \ref{eq:matching}, we use the class acquisition function, $\mathcal{L}_{con_C}$, as the loss function for class inconsistency. To stabilize the training, we compute the localization inconsistency loss as~\cite{DBLP:conf/nips/JeongLKK19}:
\begin{small}
\begin{equation}
\begin{split}
\mathcal{L}_{con_L}(\bm{b'}_i, \bm{\hat{b}}_i) = & \frac{1}{4}(|| \bm{\delta x'}_i - (- \bm{\hat{\delta x}}_i)||^2  + || \bm{y0'}_i - \bm{\hat{y0}}_i||^2  + \\
& || \bm{w'}_i -  \bm{\hat{w}}_i||^2  + || \bm{h'}_i -  \bm{\hat{h}}_i||^2),
\end{split}
\end{equation}
\end{small}

\noindent where, as we use horizontal flipping, we apply the negation on the displacement of the center $\bm{\hat{\delta x}}_i$.

% Considering that we are basing our acquisition function in the inconsistency of the detector, we train the detector to provide consistent predictions. The intuition behind training for this task is that if the detector has high inconsistency in some image, it means that it is not able to learn from it in a self-supervised manner. Thus, the only way to learn from that image is to label it during the AL cycle. During training, we mirror the active learning procedure, and we feed to the detector an image and its augmented version. Like in AL stage, we use horizontal flips as augmentation. After matching the predictions, as described in Eq. \ref{eq:matching}, we use the class acquisition function as the loss function for class inconsistency. To stabilize the training, we follow~\cite{DBLP:conf/nips/JeongLKK19} in computing the localization inconsistency loss as:
%
We compute the total consistency loss by averaging the losses from all matched pairs of predictions:
\begin{equation}
\mathcal{L}_{con} = \E[{\mathcal{L}_{con_C}(\bm{c'}, \bm{\hat{c}})}] + \E[{\mathcal{L}_{con_L}(\bm{b'}, \bm{\hat{b}})}].
\label{equ:consist}
\end{equation}

\textbf{Pseudo-labeling loss.} Our approach only pseudo-labels those objects in an image where the detector is highly confident, leaving the rest of the image unlabeled. Using the loss described in Eq. \ref{loss:multibox} would cause problems for those predictions in the regions where there are no pseudo-labels as they would be considered false positives. We thus modify the MultiBox Loss as:
\begin{small}
\begin{align}
\begin{split}
\mathcal{L}_{conf}(\bm{c}, \bm{y}, \bm{\hat{y}}) =& - \sum_{i \in Pos} \sum_{p=1}^{|\text{classes}|}  \bm{y}_{ij}^p log (\bm{c}_i^p)  \\ 
&- \sum_{i \in Neg} log (\bm{c}_i^0) - \sum_{i \in \hat{Pos}} \sum_{p=1}^{|\text{classes}|}  \bm{\hat{y}}_{ij}^p log (\bm{c}_i^p),\nonumber
\label{eq:other_nets}
\end{split}
\end{align}
\end{small}

\noindent where $\hat{y}$ and $\hat{Pos}$ represent the indicator and the positive bounding boxes for the pseudo-labels.

\textbf{Overall Training loss.} Finally, to train the deep detection network, we aggregate the multibox, $\mathcal{L}\textit{1}$ and consistency losses as:
\begin{equation}
\mathcal{L}_{total} = \mathcal{L}_{conf} + \mathcal{L}_{con} + \mathcal{L}\textit{1},
\label{eq:initial_net}
\end{equation}

\noindent where $\mathcal{L}_{con}$ is used in all samples, while $\mathcal{L}_{conf}$ and $\mathcal{L}\textit{1}$, the smooth $L_1$ for bounding boxes, are used in the labeled and pseudo-labeled samples.

\section{Experiments}
In this section, we demonstrate the effectiveness of our approach to improve the performance of object detection. For all experiments, we report mean average precision @0.5 (\textit{mAP}) as main metric, and use two public datasets: PASCAL VOC07+12 (VOC07+12) \cite{DBLP:journals/ijcv/EveringhamGWWZ10} and MS-COCO train2014 \cite{DBLP:conf/eccv/LinMBHPRDZ14}. VOC07+12 consists of $16,551$ images for training, and $4,952$ testing images taken from VOC07 testset. MS-COCO consists of $83K$ images for training, and \textit{valset2017} contains $5,000$ images for testing.

Following \cite{DBLP:conf/cvpr/YooK19, DBLP:journals/corr/abs-2103-16130}, on VOC07+12 we start by randomly sampling $2,000$ images. On the larger MS-COCO, we start by randomly sampling $5,000$ images. We perform $5$ active learning cycles, and in each cycle, we choose $1,000$ images to label. To ensure that the network does not diverge, we define each mini-batch to have half the images labeled. We set the confidence threshold for the pseudo-label threshold to $\tau = 0.99$ based on the results of the zeroth active learning step in VOC07+12.

For a fair comparison with~\cite{DBLP:conf/iclr/SenerS18, DBLP:conf/cvpr/YooK19, DBLP:conf/eccv/Agarwal0AA20, DBLP:journals/corr/abs-2103-16130}, we use the Single-Shot Detector 300 (SSD300) \cite{liu2016ssd} based on a VGG \cite{vgg} backbone for all our experiments. We train the model for $120,000$ iterations using SGD with momentum. We set the initial learning rate to $0.001$ and divide it by $10$ after $80,000$ and $100,000$ iterations, respectively. We use batches of size $32$ and a constant L2 regularization parameter set to $0.0005$. We use the same model, hyperparameters, and the same public implementation\footnote{\url{https://github.com/amdegroot/ssd.pytorch}}. We train all networks using four NVIDIA V100 GPUs. In all experiments, we train three independent networks using the same initial split of randomly sampled images and report the mean. We give the exact numbers of the mean and standard deviation in the supplementary material.

\begin{figure*}[t]
    \centering
    % \resizebox{0.99\textwidth}{!}{
    % \begin{tabular}{cccc}
    % \hspace{-0.2cm}\includegraphics[width=0.33\linewidth]{images/VOC_AL.png}&
    % \hspace{-0.2cm}\includegraphics[width=0.33\linewidth]{images/VOC_SSL.png}&
    % \hspace{-0.2cm}\includegraphics[width=0.33\linewidth]{images/VOC_ablation.png}&
    % \\(a)&(b)&(c)
    % \end{tabular}
    % }
    % \caption{\textbf{VOC07+12}: a) Comparison to state-of-the-art active learning methods; b) comparison to the two SSL method used in this work; c) ablation study on the effect of entropy, inconsistency, unified score without pseudo-labeling, and our method. * denotes ensemble method; ** denotes mixture of SSD.}
    %\includegraphics[width=0.31\linewidth]{images/VOC_AL.png}\hspace{-3mm}
    \includegraphics[width=0.31\linewidth]{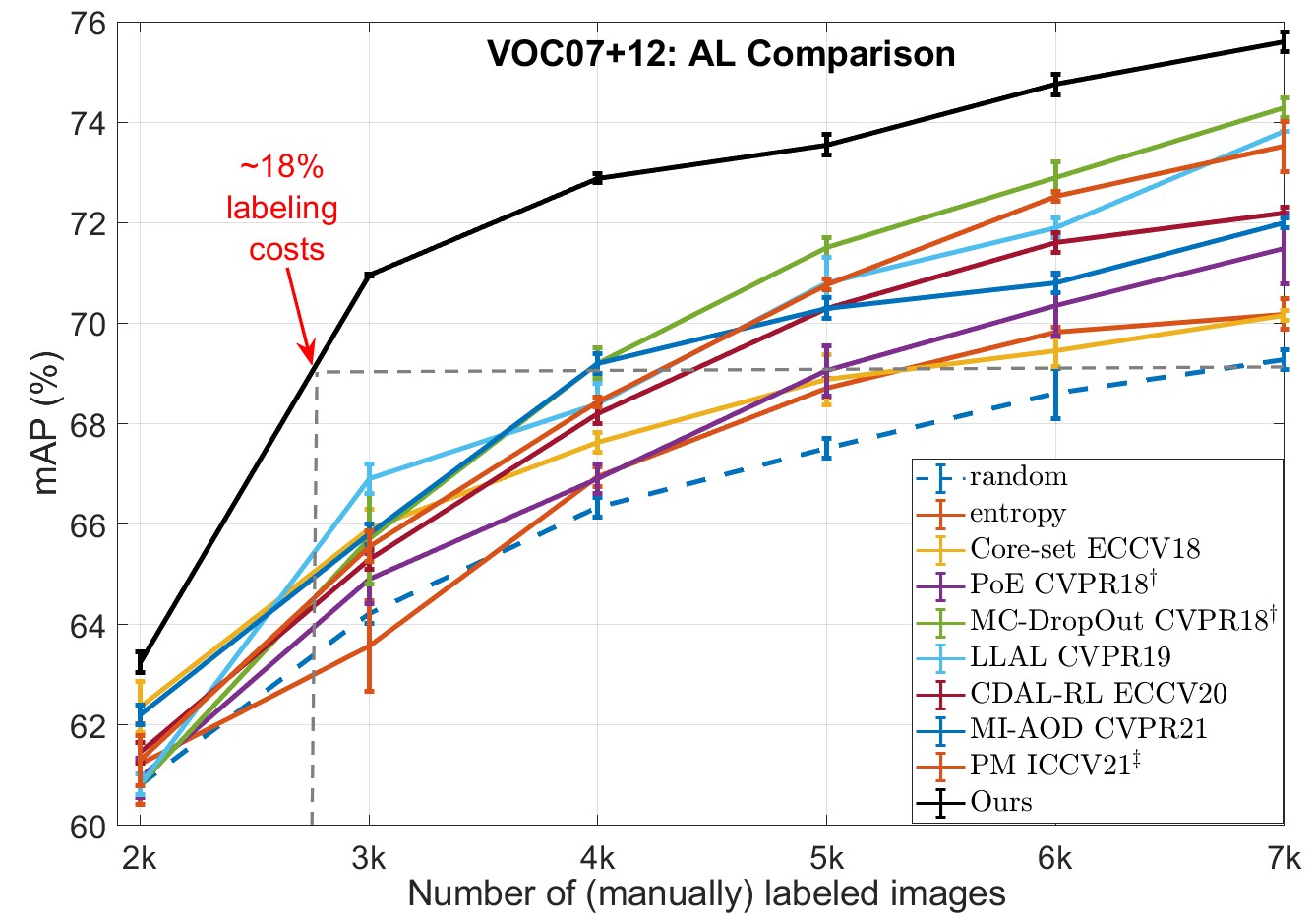}\hspace{-0.3mm}
    \includegraphics[width=0.31\linewidth]{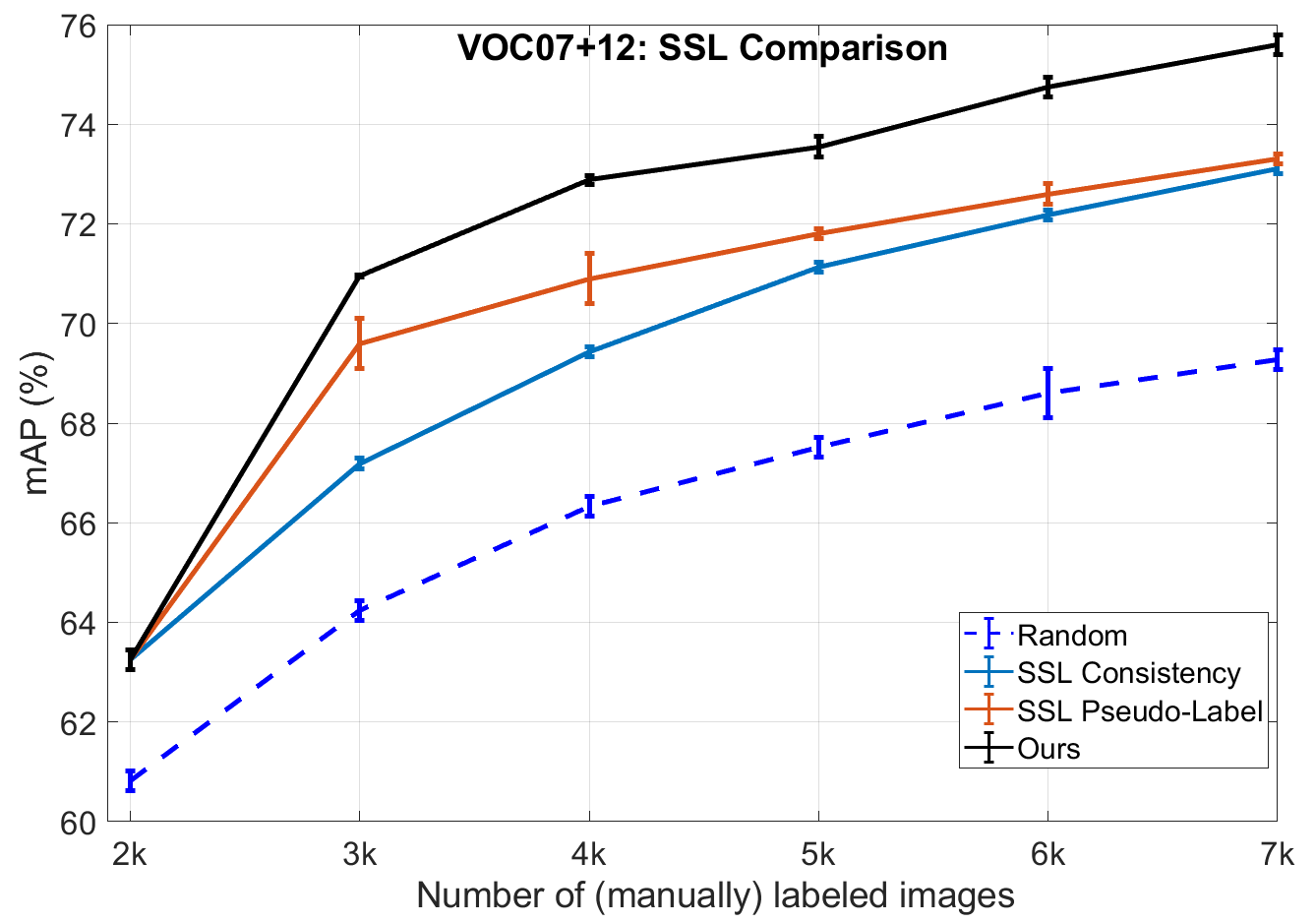}\hspace{-0.3mm}
    \includegraphics[width=0.31\linewidth]{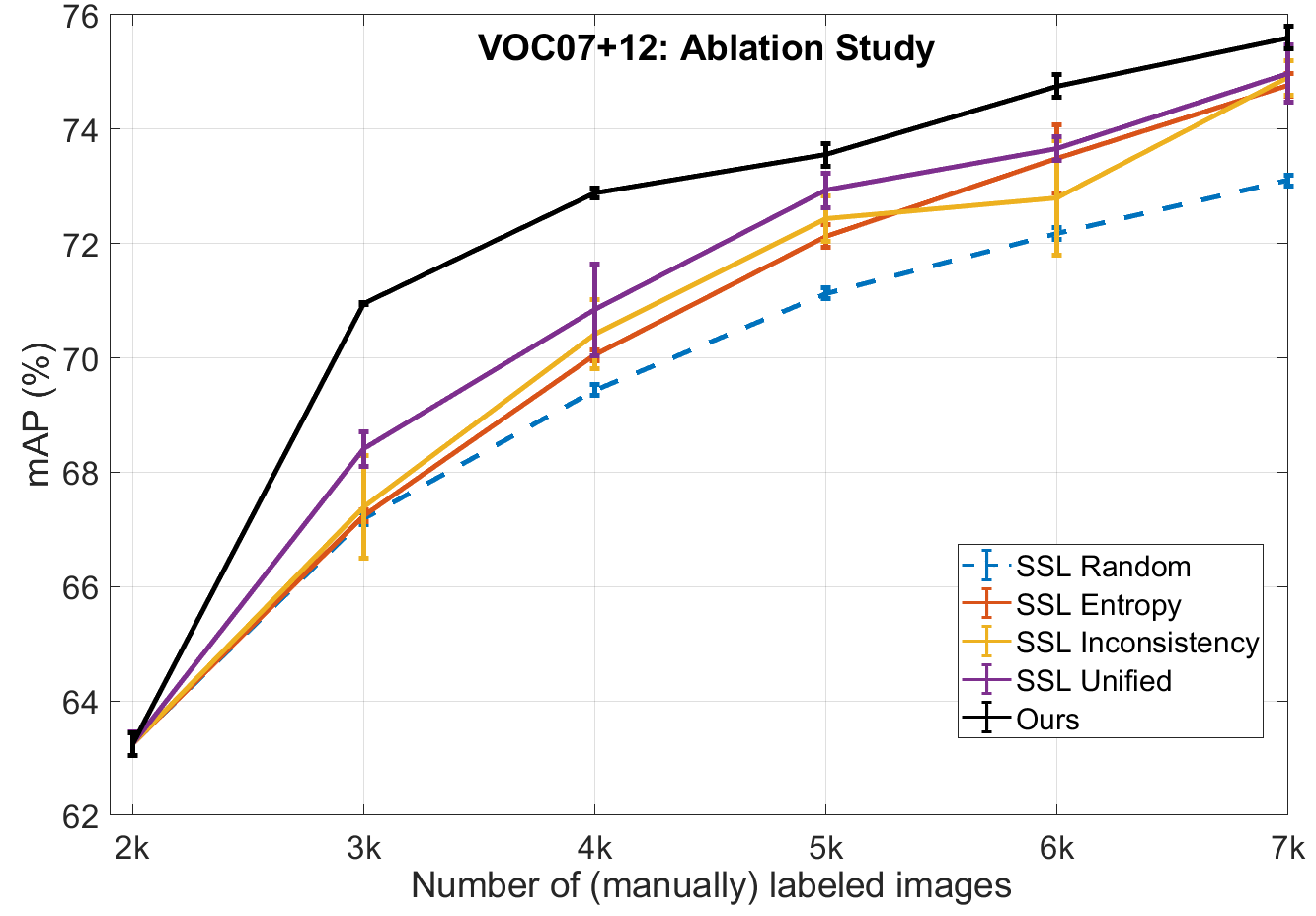}\hspace{-0.3mm}
    \caption{\textbf{VOC07+12}. \textbf{Left}: Comparison to state-of-the-art active learning methods; \textbf{Middle}: Comparison to the two SSL methods used in this work when they do not use AL; \textbf{Right}: Ablation study on the effect of entropy, inconsistency, unified score without pseudo-labeling, and our method. $\dagger$ denotes ensemble method; $\ddagger$ denotes mixture of SSD.}
    \label{fig:results_voc}
\end{figure*}

\begin{figure*}[t]
    \centering
    % \resizebox{0.99\textwidth}{!}{
    % \begin{tabular}{cccc}
    % \hspace{-0.2cm}\includegraphics[width=0.33\linewidth]{images/MS-COCO_AL.png}&
    % \hspace{-0.2cm}\includegraphics[width=0.33\linewidth]{images/MS-COCO_SSL.png}&
    % \hspace{-0.2cm}\includegraphics[width=0.33\linewidth]{images/MS-COCO_ablation.png}&
    % % \\(a)&(b)&(c)
    % \end{tabular}
    % }
    \includegraphics[width=0.31\linewidth]{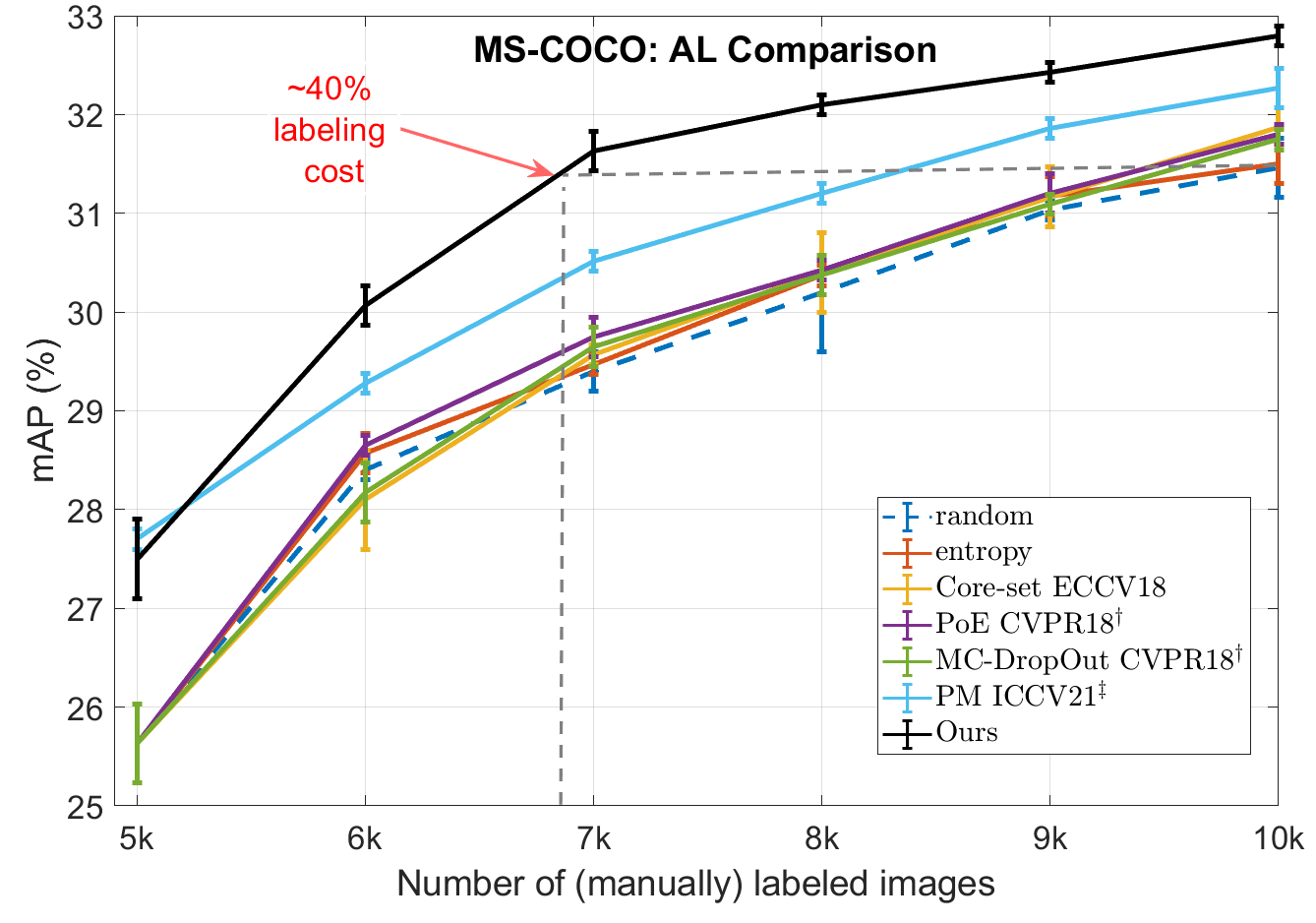}\hspace{-0.3mm}
     \includegraphics[width=0.31\linewidth]{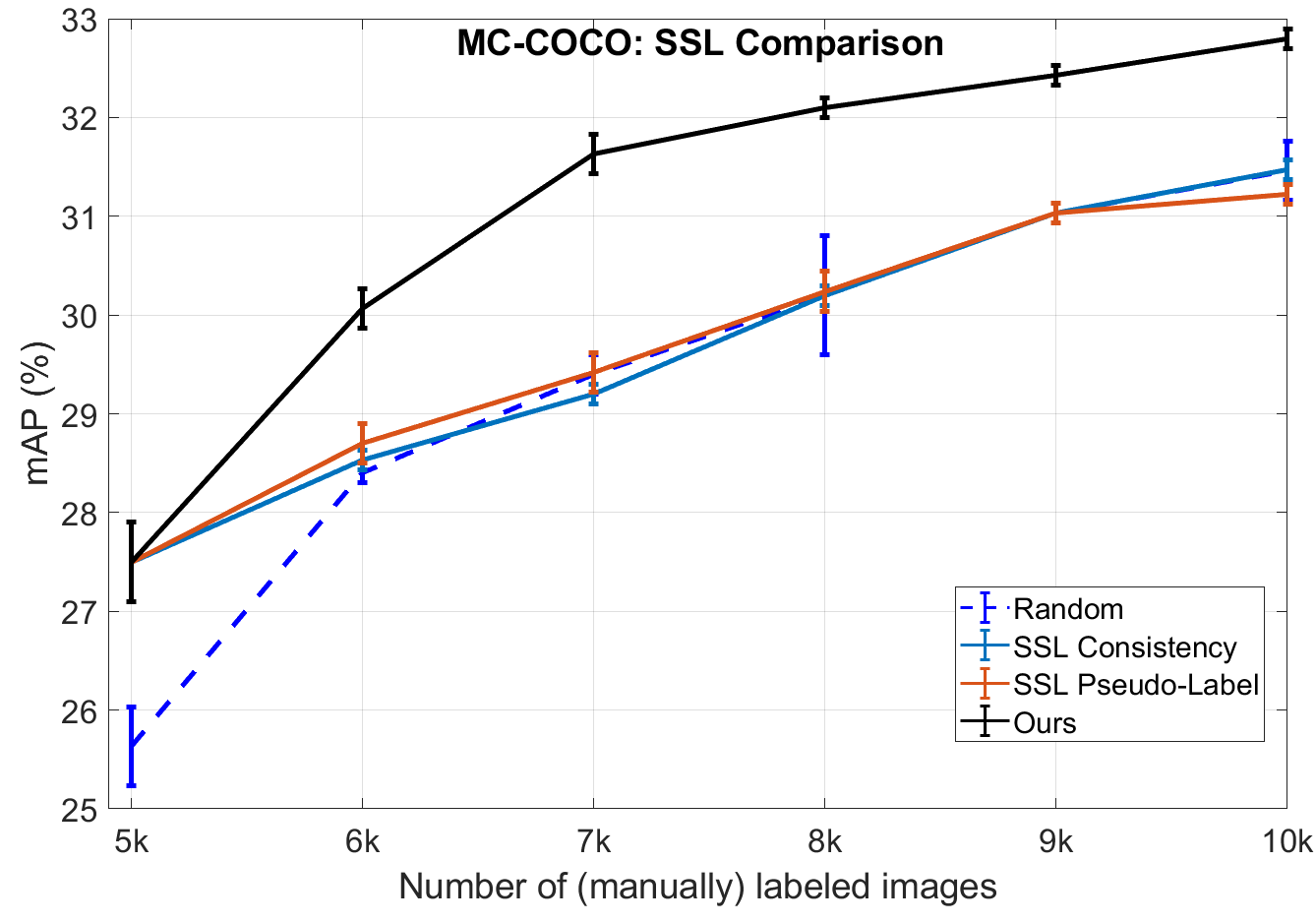}\hspace{-0.3mm}
    \includegraphics[width=0.31\linewidth]{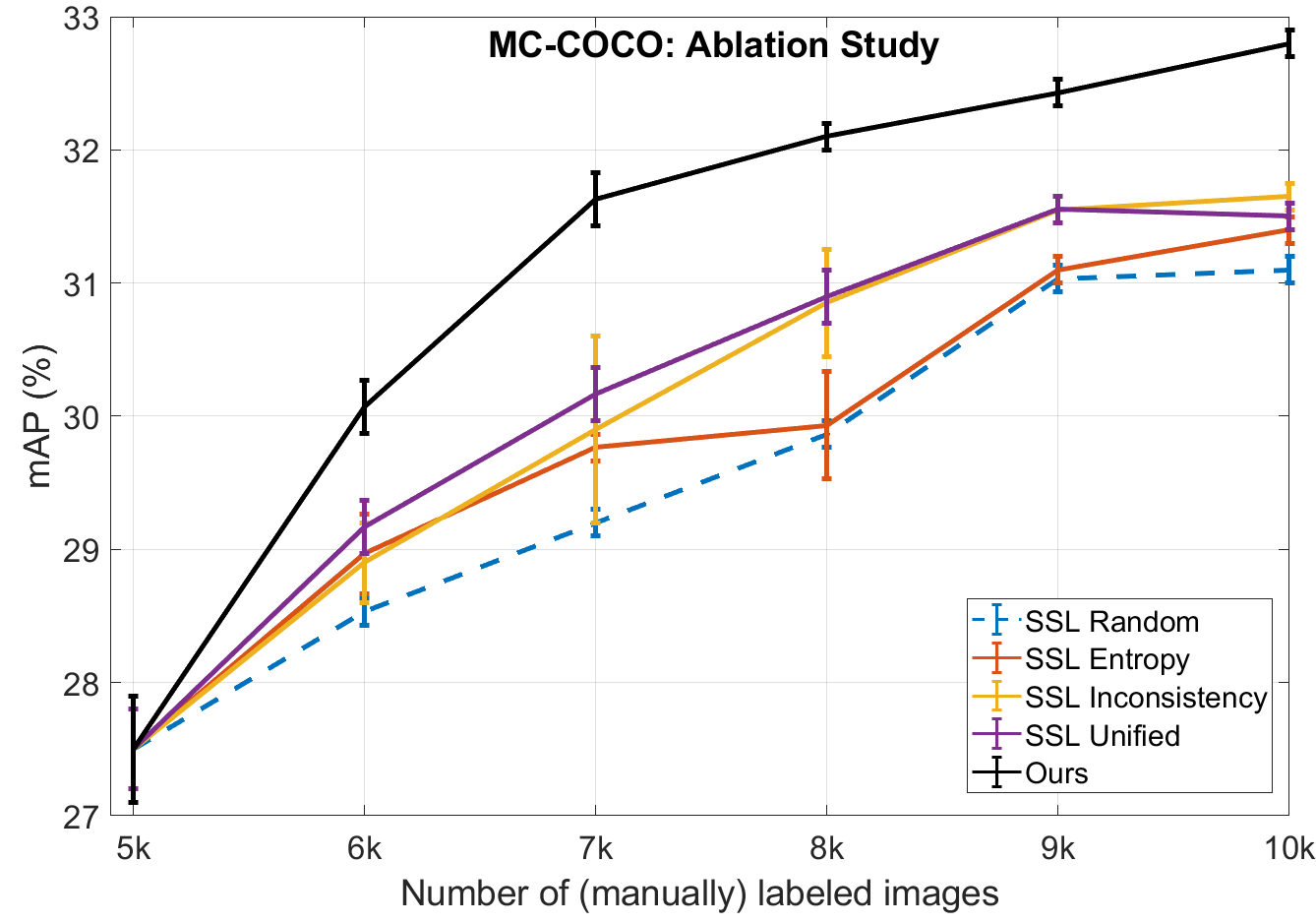}\hspace{-0.3mm}
    \caption{\textbf{MS-COCO}. \textbf{Left}: Comparison to state-of-the-art active learning methods; \textbf{Middle}: Comparison to the two SSL methods used in this work when they do not use AL; \textbf{Right}: Ablation study on the effect of entropy, inconsistency, unified score without pseudo-labeling, and our method. 
    $\dagger$ denotes ensemble method; $\ddagger$ denotes mixture of SSD.}
    \label{fig:results_coco}
    \vspace{-0.3cm}
\end{figure*}

\subsection{Main results: Comparison with other methods}
We compare our method with two baselines, random and entropy sampling, in addition to five state-of-the-art single-model methods:
Coreset~ \cite{DBLP:conf/iclr/SenerS18}, Learning Loss \cite{DBLP:conf/cvpr/YooK19}, CDAL-AL \cite{DBLP:conf/eccv/Agarwal0AA20}, MI-ALD \cite{DBLP:conf/cvpr/YuanWFLXJY21} and PM \cite{DBLP:journals/corr/abs-2103-16130}. The last method uses a mixture of SSD, thus adds extra parameters. We also compare with two multimodel approaches, MC-dropout \cite{DBLP:conf/icml/GalIG17} and ensemble-based \cite{DBLP:conf/cvpr/BeluchGNK18} active learning (consisting of three neural networks). Finally, we compare to the consistency-based SSL method and to a pseudo-labeling method.

We present the comparisons with AL methods for VOC07+12 in Fig.~\ref{fig:results_voc}a. We observe that in the first active learning cycle, our method has a relative improvement over the random baseline by $10.5\%$, and over $8.2\%$ compared to the best overall active learning method \cite{DBLP:conf/cvpr/YuanWFLXJY21}. 
We see that the performance improvement of our method is maintained in the other active learning cycles. In the last one, where we use $7,000$ samples, $5,000$ of which are actively labeled, our method outperforms the random baseline by $9.1\%$ and the best existing active learning methods by more than $2.8\%$ \cite{DBLP:journals/corr/abs-2103-16130}.
Multi-model active learning networks, namely, ensemble \cite{DBLP:conf/cvpr/BeluchGNK18} or MC-dropout \cite{DBLP:conf/icml/GalIG17} outperform single models at the cost of longer training and active learning time, and in the case of the ensemble has $3$ times more training parameters. Nonetheless, our proposed single model still reaches better results than multi-model methods, outperforming the ensembles by $8\%$ in the first AL cycle, and $1.8\%$ in the last cycle.
In Fig.~\ref{fig:results_voc}b. we compare the results of our method, with the two semi-supervised learning methods. In the first AL cycle, our method outperforms the consistency-based SSL by $5.6\%$, and the pseudo-labeling method by $2\%$. In the last cycle, we outperform the consistency method by $3.4\%$ and the pseudo-labeling method by more than $3\%$.

For MS-COCO, in Fig.~\ref{fig:results_coco}a-b, we observe that in the first active learning cycle, our method outperforms the random baseline by $5.8\%$, the best-performing AL method by $2.7\%$ \cite{DBLP:journals/corr/abs-2103-16130}, the semi-supervised method by $5.4\%$, and the ensembles by $5\%$. In the second cycle, our approach outperforms all the other methods, including PM \cite{DBLP:journals/corr/abs-2103-16130}, by almost $4\%$ or more. We observe that this difference is maintained in the other cycles, including the last active learning cycle where our method outperforms the semi-supervised method, multi-model methods \cite{DBLP:conf/cvpr/BeluchGNK18, DBLP:conf/icml/GalIG17} and the best AL method \cite{DBLP:journals/corr/abs-2103-16130} by $1.6\%$.

\subsubsection{Ablation study.}

\begin{figure*}[t]
    \centering
    \resizebox{.99\textwidth}{!}{
    \begin{tabular}{ccccc}
    \includegraphics[width=0.225\linewidth]{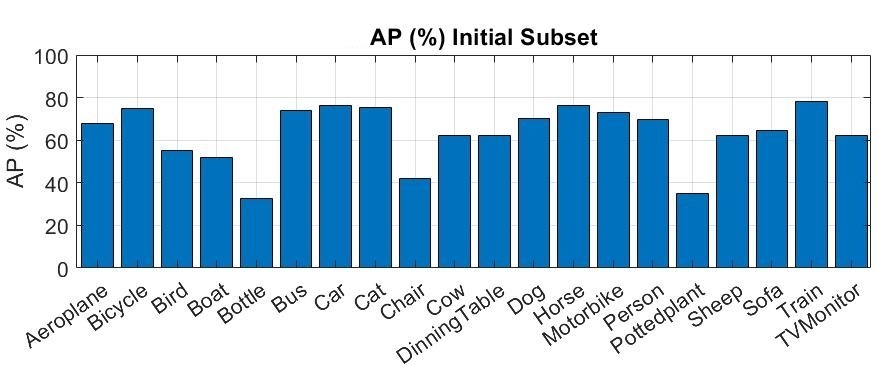}&\hspace{-0.1cm}\includegraphics[width=0.15\linewidth]{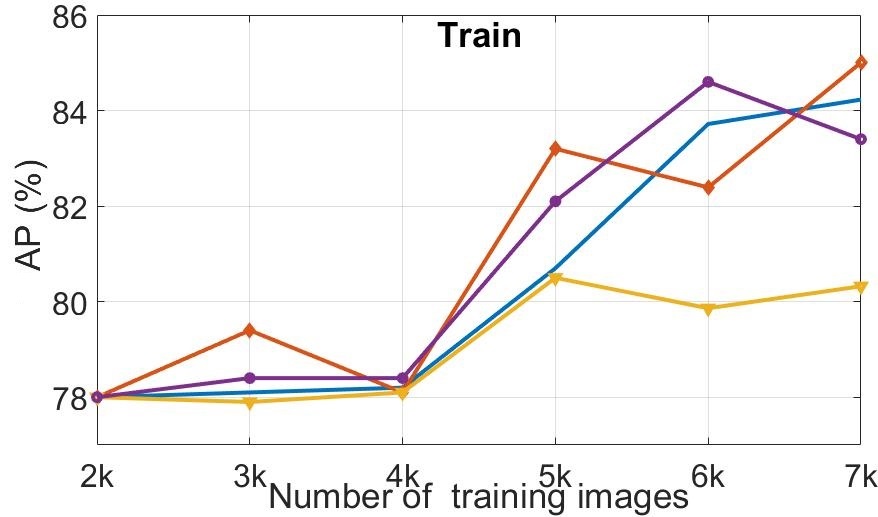}&
    \hspace{-0.2cm}\includegraphics[width=0.15\linewidth]{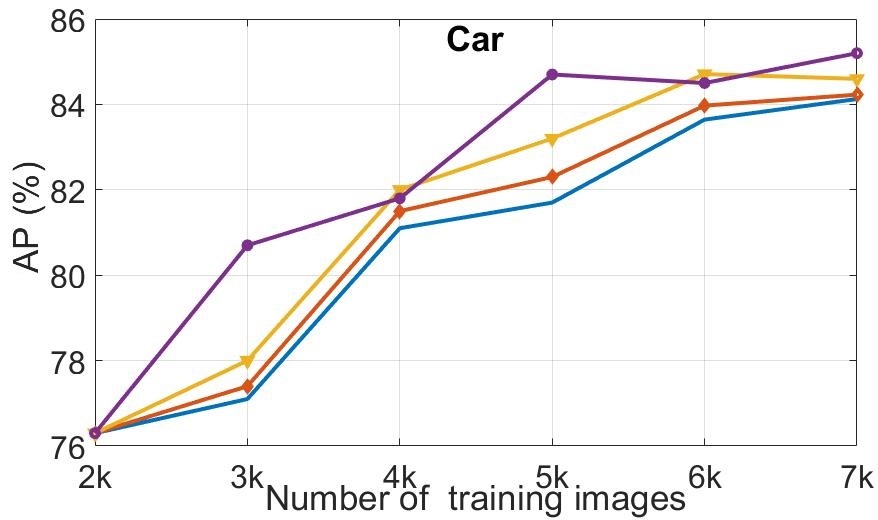}&
    \hspace{-0.2cm}\includegraphics[width=0.15\linewidth]{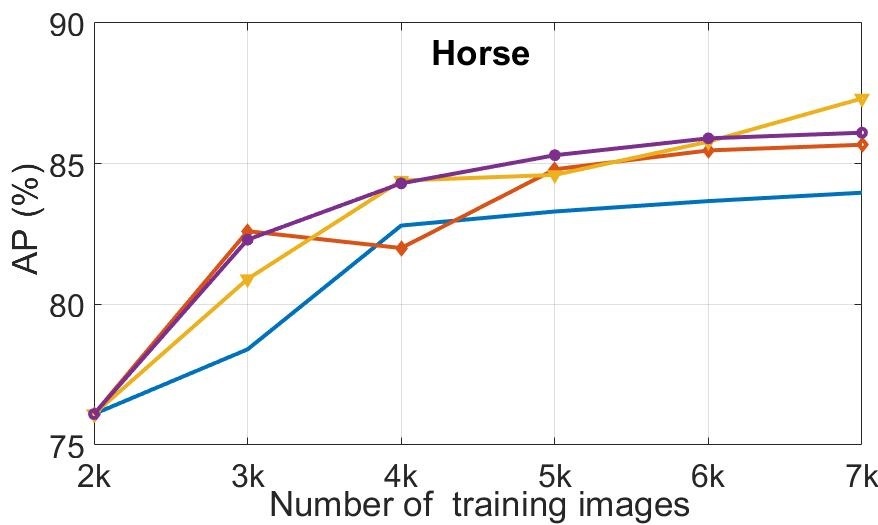}&{\hspace{-0.1cm}\includegraphics[width=0.0825\linewidth]{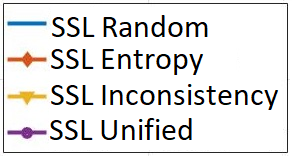}}\\
    \includegraphics[width=0.225\linewidth]{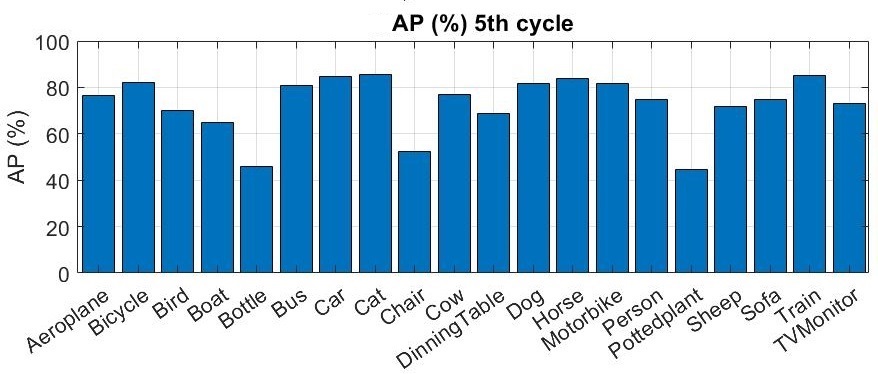}&\hspace{-0.1cm}\includegraphics[width=0.15\linewidth]{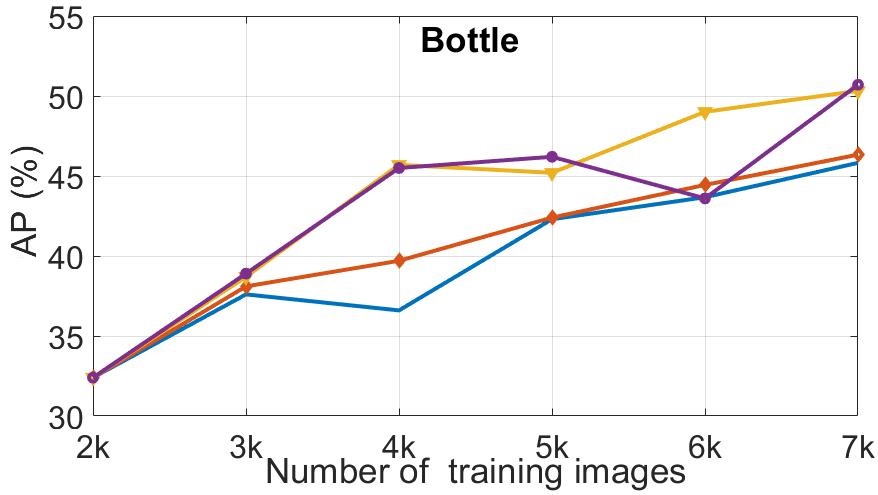}&
    \hspace{-0.2cm}\includegraphics[width=0.15\linewidth]{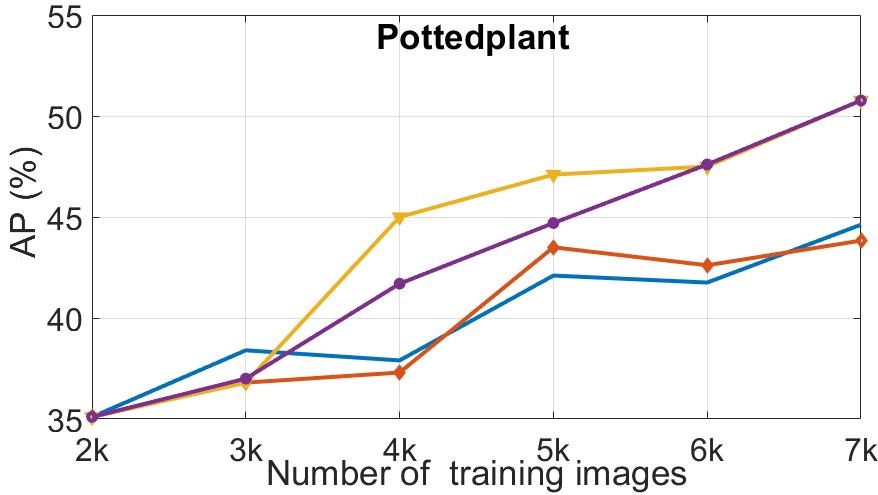}&
    \hspace{-0.2cm}\includegraphics[width=0.15\linewidth]{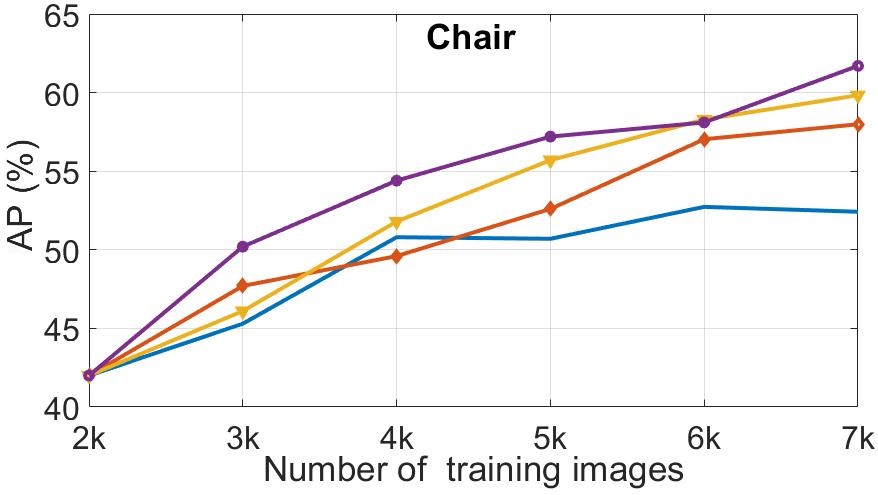}&\\
    \end{tabular}
    }
    \caption{\textbf{VOC07+12.} In the bar plots we show the accuracy per class using random sampling in the zeroth and last cycle. We present the results of each AL method for the three best-performing (\textit{"Train"}, \textit{"Car"}, and \textit{"Horse"}) and worst-performing (\textit{"Bottle"}, \textit{"Pottedplant"}, and \textit{"Chair"}) classes.} 
    \label{fig:easy_hard_classes}
    \vspace{-0.3cm}
\end{figure*}

\begin{figure}[!t]
    \centering
    \begin{tabular}{cc}
    \hspace{-0.5cm}\includegraphics[width=0.5\linewidth]{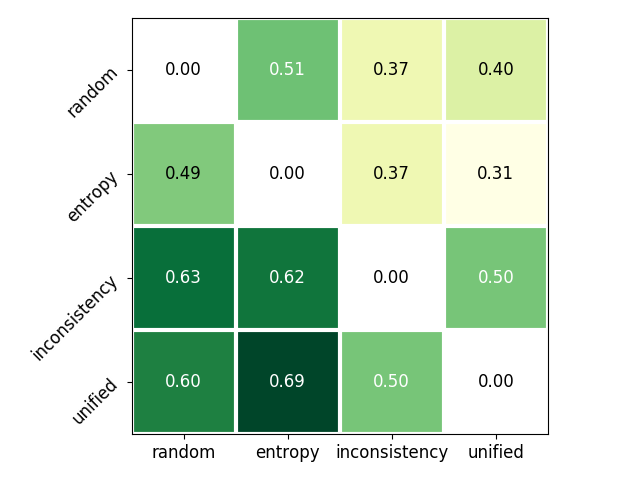}&
    \hspace{-0.7cm}    \includegraphics[width=0.5\linewidth]{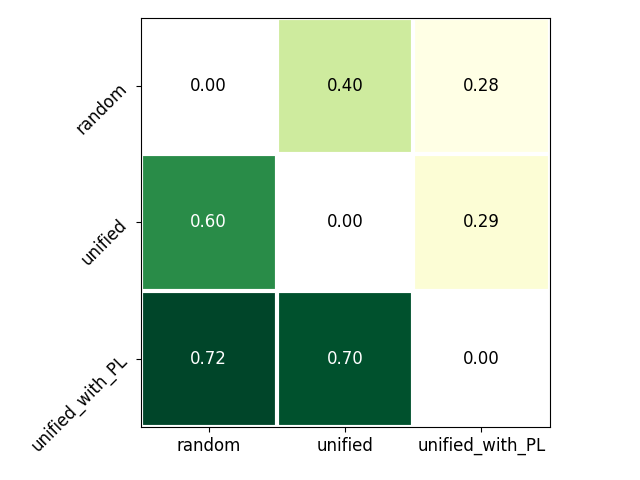}\\
    (a)&(b)
    \end{tabular}
    \caption{\textbf{MS-COCO.} a) The percentage of classes where one acquisition function outperforms another; b) The percentage of classes where our unified acquisition function outperforms random with and without pseudo-labels. Example: taking the entry "unified" in the y-axis, and "entropy" in the x-axis, we get the value $0.69$ which means that "unified" acquisition function outperforms the "entropy" acquisition function in $69\%$ of classes.}
    \label{fig:per_class_coco}
    \vspace{-0.3cm}
\end{figure}

\textbf{The effect of active learning and auto-labeling}.
We now analyze every module of our method, in order to disentangle the contribution coming from them. In Fig.~\ref{fig:results_voc}c, we present the performance comparison of the semi-supervised model on VOC07+12 under different acquisition functions (random, entropy, inconsistency) and two instances of our unified method: with and without pseudo-labels. We see that on the first active learning cycle, neither entropy nor inconsistency significantly outperforms the results of random sampling. 
However, we immediately see a significant effect, i.e., a relative improvement over $0.9\%$ using entropy and $1.4\%$ using inconsistency, in the second active learning cycle. We see that the increase in performance gets bigger in the next AL cycle and in the fifth active learning cycle, the performance gain from entropy is $2.3\%$ and from inconsistency is $2.4\%$. 

We then show the results of our unified acquisition function. In the first active learning cycle, we immediately see a significant improvement in performance. While entropy ($67.24mAP$) and inconsistency ($67.39$mAP) reach an insignificant improvement over random sampling ($67.19mAP$), our acquisition function reaches $68.40mAP$, which is $1.5\%$ better than random sampling. The performance improvement gets larger in the next cycles: $2\%$ in the second cycle, $2.5\%$ in the third cycle, and a peak improvement of $2.6\%$ in the fifth active learning cycle. In all cases, our proposed score outperforms both active learning methods that are based on a single acquisition function.

We further study the effect of pseudo-labeling in our framework. In Fig. \ref{fig:results_voc}c, we observe that on the first active learning cycle, adding pseudo-labels comes with an immediate boost, improving the results by $3.7\%$ compared to the already well-performing acquisition function for semi-supervised learning ($5.3\%$ better than the semi-supervised method that uses random sampling).  We further observe that on the second cycle, it gives an improvement of $2.9\%$ compared to using only our acquisition function ($4.7\%$ better than the semi-supervised method that uses random sampling). The Pseudo-labeling module continues to give a boost in performance in all the following AL steps.%, although we see saturation in the performance boost.

%We then evaluate our method when we exclude the outliers, the predictions for which the network's confidence is lower than $\tau_1=0.5$. We see that this improves our results by a further $1.5\%$ in the first AL cycle and by $1.4\%$ in the second AL cycle. The performance gain in the later iterations gets lower, nevertheless, we still improve the results by $0.3\%$ in the fifth AL cycle.

In Fig.~\ref{fig:results_coco}c, we provide a similar ablation study for MS-COCO. We again observe that the unified score outperforms both the entropy and inconsistency scores in isolation. However, unlike in VOC07+12, we observe that using only the entropy, the improvement is marginal over random sampling. On the other hand, we observe that inconsistency works significantly better than random sampling and entropy (we provide an explanation in the next section).
We further observe the effect of pseudo-labels. We see that in the first AL cycle, adding pseudo-labels boosts the performance by $3.1\%$ and the performance boost is maintained up to the last cycle. This is very different from the results of pseudo-labeling alone, see Fig.~\ref{fig:results_coco}b, where the performance gain is marginal. In other words, pseudo-labeling in isolation does not work well. However, pseudo-labeling complemented with the unified score reaches high results.

%why the combined acquisition function performs better than either inconsistency or entropy
\textbf{Acquisition functions.} We now focus on analyzing the effects of aggregating the two acquisition functions. In Fig. \ref{fig:easy_hard_classes}, we check the performance of every individual class in the zeroth and the last AL cycle in VOC07+12 dataset. We then focus on three best-performing classes (\textit{"Train"}, \textit{"Car"}, and \textit{"Horse"}) and three worst-performing classes (\textit{"Bottle"}, \textit{"Pottedplant"}, and \textit{"Chair"}). A first observation is that for the best-performing classes, entropy-based AL, on average, tends to outperform inconsistency-based AL. %We also see that while the inconsistency score does a good job in classes \textit{"Car"} and \textit{"Horse"}, it entirely fails on the best performing class \textit{"Train"}, performing worse than the random sampling.

On the other hand, we see that inconsistency-based AL outperforms the entropy-based AL by a significant margin in all three worst-performing classes. While the entropy-based AL on average seems to only slightly outperform random sampling, the inconsistency-based AL gives a relative performance gain of up to $24\%$, $14\%$ and $18\%$ in classes \textit{"Bottle"}, \textit{"Pottedplant"}, and \textit{"Chair"}. Intuitively, one can argue that this phenomenon is to be expected. The fact that the network does a poor job on its predictions leads to its class predictions being unreliable for any uncertainty-based AL method. At the same time, a more general acquisition function dependent only on the robustness of the network is better suited for low-performing classes. Finally, we show that our acquisition function reaches the best overall results.

Because of the massive number of classes, we aggregate the results on MS-COCO dataset. In Fig. \ref{fig:per_class_coco}a, we show the percentage of classes where one acquisition function outperforms the other. We see that inconsistency outperforms entropy in $62\%$ of the classes, and our unified score outperforms entropy in $60\%$ of the classes. This explains why in MS-COCO, which contains many more challenging classes, the robustness-based acquisition scores significantly outperform the uncertainty-based acquisition score. We provide the results for each AL cycle in the supplementary.

\textbf{Do we need SSL training?} The inconsistency acquisition function computes the robustness of the network for each image in the acquisition pool. If the network is inconsistent in an image, despite that it was trained to minimize the inconsistency of that image, then that image provides information that was not captured by the SSL. We now check what happens if the network is not trained in SSL, thus it has never seen the images in the labeling pool. In Tab. \ref{tab:inconsistency_trained}a, we show the results of inconsistency acquisition function for a network trained with and without consistency loss. As expected, the results with the SSL loss significantly outperform the results of the fully-supervised baseline. Interestingly, the results of inconsistency AL are not better than those of random. Clearly, in order to be able to exploit the robustness information in samples, the network needs to try minimizing their inconsistency during training. 

% \setlength{\columnsep}{9pt}
% \begin{wraptable}{r}{0.23\textwidth}
% \vspace{-3mm}
% \centering
% \resizebox{\linewidth}{!}{%
% \begin{tabular}{c|c cc}
% \toprule
% \textbf{Cycle} & \textbf{Random} &\textbf{No SSL} & \textbf{SSL}\\ \hline
% 1 & 64.23 & 63.26 & \textbf{67.39}  \\
% 2 & 66.33 & 65.79 & \textbf{70.42}  \\
% 3 & 67.51 & 67.16 & \textbf{72.43}  \\
% 4 & 68.60 & 68.65 & \textbf{72.80}  \\
% 5 & 69.27 & & 74.90  \\
% \end{tabular}%
% }
% \caption{\textbf{VOC07+12:} The effect of training on SSL.}
% \vspace{-2mm}
% \label{tab:inconsistency_trained}
% \end{wraptable}

\begin{figure}[!t]
    \centering
    \includegraphics[width=0.99\linewidth]{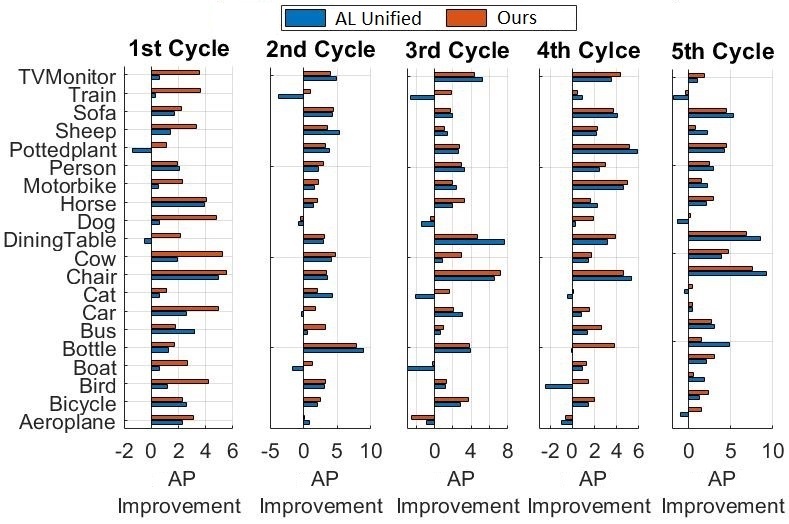}
    \caption{\textbf{VOC07+12:} Effect of pseudo-labels compared to AL alone for every class.}
    \label{fig:pseudo-labels-ration}
\end{figure}

\textbf{PL performance boost per class.}
We now study if the pseudo-labels help only some particular classes, or if they help in all classes. We start the analysis on VOC07+12 dataset. In Fig. \ref{fig:pseudo-labels-ration}, we plot the performance gain coming from the module for each class and compare it with the performance gain from AL alone, and random sampling. In the first AL cycle, we see that pseudo-labels improve over random sampling in all $20$ classes, with AL alone giving a negative boost in two classes: \textit{"Pottedplant"} and \textit{"DiningTable"}. %Furthermore, they outperform AL alone in $17$ out of $20$ classes.
We also find out that pseudo-labels give a boost over AL alone in all three worst-performing classes (\textit{"Bottle"}, \textit{"Pottedplant"}, and \textit{"Chair"}). We see a similar pattern in the other cycles. In the second cycle, the pseudo-labels module improves over AL alone in $14$ classes and gives a negative boost in only one class (\textit{"Dog"}) compared to AL alone that gives a negative boost in four classes. In the third cycle, the pseudo-labels module improves over AL alone in $11$ classes and gives a negative boost in two classes (compared to AL in five classes). In the fourth cycle, the pseudo-labels module improves over AL alone in $15$ classes and gives a negative boost in one class (compared to AL in three classes). Finally, in the fifth AL cycle, the pseudo-labels module improves over AL alone in $10$ classes with class \textit{"Car"} being a tie and gives a negative boost in only one class, compared to AL in three classes. We thus conclude that by focusing only on the \textit{hard} samples, AL alone harms the performance on several classes. However, adding pseudo-labels diminishes this effect, making the network much more robust and thus preventing a dataset drift.

We do a similar analysis on the larger MS-COCO, by aggregating the results, showing the results in Fig. \ref{fig:per_class_coco}b. While our unified acquisition function outperforms the random acquisition in $60\%$ of the classes, adding pseudo-labels increases the number of classes this happens to $72\%$, showing the usefulness of pseudo-labels. We provide the experiment results for each AL cycle in the supplementary material.

\textbf{Ratio of pseudo-labels.}
We now study the effect of increasing the number of pseudo-labels by allowing more noisy pseudo-labels. To do so, we lower the pseudo-labeling threshold $\tau$ from $0.99$ to $0.9$ and $0.5$. We present the results in Fig. \ref{fig:tau}a. We observe that we reach the best overall results by using an extremely high threshold $\tau=0.99$. Decreasing $\tau$ to $0.9$ and thus allowing more pseudo-labels harms the performance. Further decreasing it to $0.5$, hence allowing many more pseudo-labels, actually harms the entire training. We thus conclude that we need to be selective in the choice of pseudo-labels. 

To understand why the performance improvement of the network trained with the pseudo-labels module diminishes in the later active learning cycles, we study the pseudo-labels gain as a function of the pseudo-labels ratio to the entire labels. As we show in Fig. \ref{fig:tau}b, in the first active learning cycle where the pseudo-labels bring a maximum gain ($3.7\%$), roughly half of the labels are pseudo-labels. With the decrease of the number of pseudo-labels, we see a tendency for the gain to lower. Our intuition is that when the number of pseudo-labels is high, despite them being noisy, they still help the training process. An interesting fact is that while the total number of pseudo-labels decreases for each AL cycle (because $1,000$ images are removed from the labeling pool) the number of pseudo-labels for an image increases with each cycle, starting from $0.58$ in the first cycle, to $0.81$ in the last one. Thus, the network becomes better at selecting pseudo-labels during AL cycles. % However, when the number of pseudo-labels gets lower, their effect gets smaller. 

On MS-COCO, where the number of images that can be potentially pseudo-labeled is higher ($78K$ compared to $14,651$ in VOC0712), the performance gain from the pseudo-labels module does not diminish. This is because the ratio of pseudo-labels in all cycles remains high. 

\begin{table}[!t]
\centering
\begin{tabular}{cc}
\resizebox{0.5\linewidth}{!}{%
\begin{tabular}{c|c cc}
\toprule
\textbf{Cycle} & \textbf{Random} &\textbf{No SSL} & \textbf{SSL}\\ \hline
1 & 64.23 & 63.26 & \textbf{67.39}  \\
2 & 66.33 & 65.79 & \textbf{70.42}  \\
3 & 67.51 & 67.16 & \textbf{72.43}  \\
4 & 68.60 & 68.65 & \textbf{72.80}  \\
5 & 69.27 & 70.33 & \textbf{74.90}  \\
\end{tabular}%
}
&
\resizebox{0.43\linewidth}{!}{%
\begin{tabular}{c|ccc}
\toprule
\textbf{Cycle} & \textbf{0.5} & \textbf{0.9} & \textbf{0.99} \\ \hline
1 & 80.04 & 91.13 & \textbf{96.68} \\
2 & 84.00 & 92.21 & \textbf{96.01} \\
3 & 86.00 & 93.32 & \textbf{95.80} \\
4 & 87.55 & 93.41 & \textbf{95.61} \\
5 & 90.05 & 94.64 & \textbf{95.57} \\
\end{tabular}%
}
\\(a)&(b)
\end{tabular}
\vspace{-3.5mm}
\caption{\textbf{VOC07+12.} a) The effect of training on SSL. b)  PL correctness with $\tau$.}
\vspace{-5mm}
\label{tab:inconsistency_trained}
\end{table}

\iffalse
\begin{figure}[!t]
    \centering
    \begin{tabular}{cc}
    \hspace{-2.8mm}\includegraphics[width=0.885\linewidth]{images/tau2.jpg}\\
    \hspace{1mm}\includegraphics[width=0.85\linewidth]{images/pseudoLabels.jpg}\\
    % (a)&(b)
    \end{tabular}
    \caption{\textbf{VOC07+12.} \textbf{Top}: Accuracy as a function of $\tau$ for selecting pseudo-labels. \textbf{Bottom}: Accuracy improvement with respect to the pseudo-labels ratio to the entire labels.}
    \label{fig:tau}
\end{figure}
\fi

\begin{figure}[!t]
	\centering
	\begin{tabular}{cc}
	\hspace{-0.25cm}\includegraphics[width=0.5\linewidth]{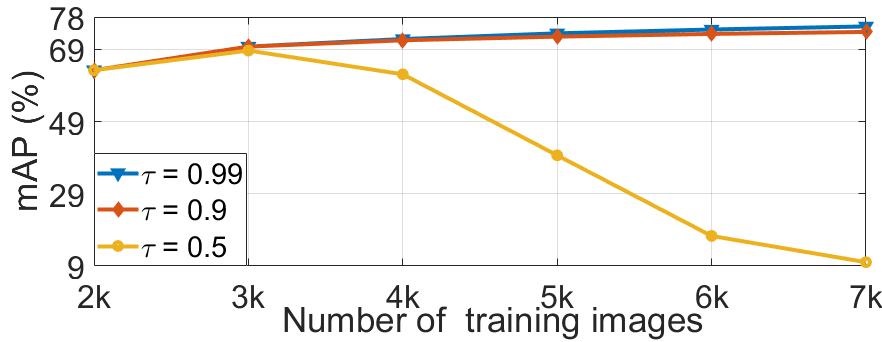}&
    \hspace{-0.5cm}	\includegraphics[width=0.5\linewidth]{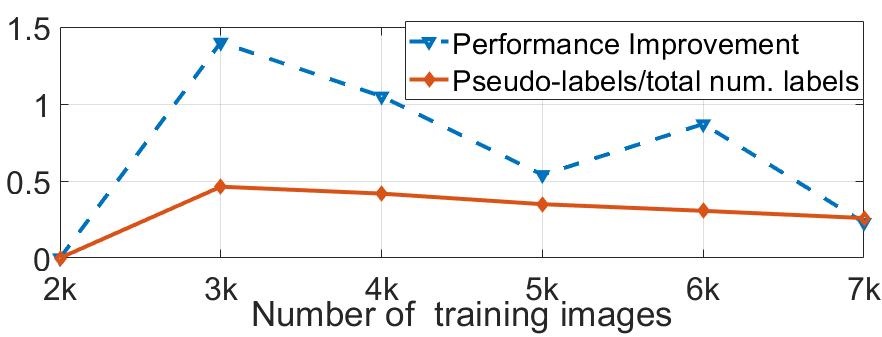}\\
    (a)&(b)
	\end{tabular}
	%\vspace{-0.5cm}
    \caption{\textbf{VOC07+12.} \textbf{Left}: Accuracy as a function of $\tau$ for selecting pseudo-labels. \textbf{Right}: Accuracy improvement with respect to the pseudo-labels ratio to the entire labels.}
    \label{fig:tau}
\end{figure}

\textbf{Pseudo-labels noise.} Deep neural networks are overconfident and not well-calibrated, thus inducing pseudo-labeling errors.  We consider a pseudo-label correct if the predicted class is the same as the ground truth and intersection over union with the ground truth object is over $0.5$. 
We provide the correctness of the pseudo-labels given by our model in Tab.~\ref{tab:inconsistency_trained}b. We see that setting the pseudo-labeling threshold to $0.99$ leads to $3.7\%$ pseudo-labeling errors. This percentage remains constant over different AL cycles suggesting the network is robust to this amount of noise.

% \textbf{Limitations.} Our method is task-specific and limited to a set of known classes and has not been tested in an open-world setting. Furthermore, our method is task-specific and thus less suitable for acquiring datasets for multi-task networks. Finally, we only experimented with a vanilla pseudo-labeling method, which might not reach the best possible results.
\textbf{Limitations.} Our method is task-specific and limited to a set of known categories. We did not conduct any experiment on multi-task networks or in an open-world setting. %Furthermore, our method is task-specific and thus less suitable for acquiring datasets for multi-task networks. Finally, we only experimented with a vanilla pseudo-labeling method, which might not reach the best possible results.

\section{Conclusions}

In this work, we developed a framework that reduces the labeling costs for object detection. Our framework consists of a novel acquisition function based on the \textit{robustness} of the neural network with respect to its predictions, and an auto-labeling scheme that prevents a potential distribution drift. In this way, our unified model chooses to actively label the most informative samples in the dataset, while it pseudo-labels the easiest samples. This allows us to use the majority of the dataset in a supervised manner while reducing the labeling costs. As we showed in the experiments, we can reduce the labeling costs by up to $82\%$ in order to reach the same results as a fully-supervised baseline.  

%%%%%%%%% REFERENCES
{\small
\bibliographystyle{unsrt}
\bibliography{egbib}
}

\clearpage

\twocolumn[{%
\renewcommand\twocolumn[1][]{#1}%
\vspace{-3em}
\begin{center}
    \centering
    \Large{\textbf{--- Supplementary material ---}}
    
\end{center}%
\vspace{1em}
}]

\setcounter{section}{0}

\section{Exact numbers for the experiments given in the main paper}

In the paper, we provide plots for the main experiments due to the limited space. In Tables~\ref{voc-results-a}, \ref{tab:2}a, \ref{tab:2}b, \ref{coco-results-a}, \ref{tab:4}a, \ref{tab:4}b, we summarize the exact numbers corresponding to Figures 3a, 3b, 3c, 4a, 4b and 4c of the main paper. We provide the mean and the standard deviation for each method and AL cycle. Each experiment has been run three times.

\begin{table*}[!h]
\centering
\resizebox{0.99\textwidth}{!}{
\begin{tabular}{l|llllllllll}
\toprule
Cycle & Random                         & Entropy                        & Core-Set \cite{DBLP:conf/iclr/SenerS18}      & LLAL \cite{DBLP:conf/cvpr/YooK19}  & Ensemble \cite{DBLP:conf/cvpr/BeluchGNK18} & MC-dropout \cite{DBLP:conf/icml/GalIG17}  & CDAL-RL \cite{DBLP:conf/eccv/Agarwal0AA20} & MI-AOD \cite{DBLP:conf/cvpr/YuanWFLXJY21} & PM \cite{DBLP:journals/corr/abs-2103-16130} &Ours                                                           \\ \hline
0    & 60.82$\pm$0.2                                          & 61.23$\pm$0.8                                          & 62.36$\pm$0.5 & 60.95$\pm$0.4 & 60.82$\pm$0.2 & 60.82$\pm$0.2 & 61.45 $\pm$0.2& 62.20 $\pm$0.2  & 61.30 $\pm$0.5  & \textbf{63.25}$\pm$0.2 \\ 
1    & 64.23$\pm$0.2                                          & 63.57$\pm$0.9                                          & 65.90$\pm$0.4 & 64.91$\pm$0.5 &  65.70$\pm$0.9 & 66.90$\pm$0.3 & 65.30 $\pm$0.2 & 65.60 $\pm$0.2 & 65.56 $\pm$0.3  & \textbf{70.95} $\pm$0.1 \\ 
2    & 66.33$\pm$0.2                                          & 66.94$\pm$0.2                                          & 67.63$\pm$0.2     & 66.90$\pm$0.3  & 69.20$\pm$0.3 & 68.40$\pm$0.2 & 68.20 $\pm$0.3 & 69.25 $\pm$0.2& 68.43 $\pm$0.1 &\textbf{72.88}$\pm$0.1 \\ 
3    & 67.51$\pm$0.2                                          & 68.70$\pm$0.2                                          & 68.88$\pm$0.5     & 69.05$\pm$0.5 & 71.50$\pm$0.2 & 70.80$\pm$0.4 &  70.30 $\pm$0.2 & 70.35 $\pm$0.2 & 70.77 $\pm$0.1   & \textbf{73.55}$\pm$0.2 \\ 
4    & 68.60$\pm$0.5                                          & 69.82$\pm$0.1                                          & 69.44$\pm$0.3     & 70.35$\pm$0.6  & 72.90$\pm$0.3 & 71.90$\pm$0.5 &  71.60 $\pm$0.2 & 70.80 $\pm$0.2 & 72.52 $\pm$0.1  & \textbf{74.75}$\pm$0.2 \\ 
5    & 69.27$\pm$0.2                                          & 70.18$\pm$0.3                                          & 70.16$\pm$0.1     & 71.49$\pm$0.7 & 74.29$\pm$0.2 & 73.81$\pm$0.0 &  72.20 $\pm$0.2 & 72.00 $\pm0.2$ & 73.52 $\pm0.5$  & \textbf{75.60}$\pm0.2$ \\ 
\end{tabular}
}
\caption{\textbf{VOC07+12}. Comparison to state-of-the-art active learning methods. We initially use $2,000$ randomly sampled images and, in every other cycle, we label $1,000$ extra images. Our method outperforms all the other methods, including ensembles, by a large margin.}
\label{voc-results-a}
\end{table*}

\begin{table*}[!t]
\centering
\begin{tabular}{cc}
\resizebox{0.42\linewidth}{!}{%
\begin{tabular}{l|llll}
\toprule
Cycle & Random                         & SSL-cons. \cite{DBLP:conf/nips/JeongLKK19}                        & SSL-PL \cite{pseudo-labelling}      &  Ours                                                           \\ \hline
0    & 60.82$\pm$0.2   & \textbf{63.25}$\pm$0.2 &      \textbf{63.25}$\pm$0.2                                    & \textbf{63.25}$\pm$0.2 \\ 
1    & 64.23$\pm$0.2                                          & 67.19 $\pm$0.1& 69.60 $\pm$0.5  & \textbf{70.95} $\pm$0.1 \\ 
2    & 66.33$\pm$0.2                                          &   69.44 $\pm$0.1                                       & 70.90 $\pm$0.5  &\textbf{72.88}$\pm$0.1 \\ 
3    & 67.51$\pm$0.2                                          & 71.13 $\pm$0.1 &  71.80 $\pm$0.1  & \textbf{73.55}$\pm$0.2 \\ 
4    & 68.60$\pm$0.5                                                                                   & 72.18 $\pm$0.1& 72.60 $\pm$0.2 & \textbf{74.75}$\pm$0.2 \\ 
5    & 69.27$\pm$0.2                                          & 73.10 $\pm$0.1& 73.30 $\pm$0.2  & \textbf{75.60}$\pm$0.2 \\ 
\end{tabular}
}
&
\resizebox{0.52\linewidth}{!}{%
\begin{tabular}{l|ccccc}
\toprule
Cycle & Random  & Entropy & Inconsistency & Combined  &  Ours \\ \hline
0  &  \textbf{63.25}$\pm$0.2 & \textbf{63.25}$\pm$0.2 & \textbf{63.25}$\pm$0.2 & \textbf{63.25}$\pm$0.2 &   \textbf{63.25}$\pm$0.2                                          \\ 
1     & 67.19 $\pm$0.1 & 67.24$\pm$0.1 & 67.39$\pm$0.9 & 68.40$\pm$0.3  & \textbf{70.95} $\pm$0.1 \\ 
2     & 69.44$\pm$0.1 & 70.05$\pm$0.1 & 70.42$\pm$0.6 & 70.84$\pm$0.8 &  \textbf{72.88}$\pm$0.1 \\ 
3     & 71.13$\pm$0.1 & 72.13$\pm$0.2 & 72.43$\pm$0.4 & 72.93$\pm$0.3 &  \textbf{73.55}$\pm$0.2 \\ 
4     & 72.18$\pm$0.1 & 73.48$\pm$0.6 & 72.80$\pm$1.0 & 73.66$\pm$0.2 & \textbf{74.75}$\pm$0.2 \\ 
5     & 73.1$\pm$0.1 & 74.77$\pm$0.2 & 74.90$\pm$0.3 & 74.98$\pm$0.5   & \textbf{75.60}$\pm$0.2\\ 

\end{tabular}
}
\\(a)&(b)
\end{tabular}
\caption{\textbf{VOC07+12}. a) Comparison to two semi-supervised learning methods. We initially use $2,000$ randomly sampled images and, in every other cycle, we label $1,000$ extra images. Our method outperforms both of them by a large margin. b) Ablation study on the effect of entropy, inconsistency, unified score, and our method in VOC07+12. We observe that doing active learning with either entropy or consistency outperforms the semi-supervised model, that the unified score performs better than either of the individual scores, and that our method reaches the best overall results.}
\label{tab:2}
\end{table*}

\begin{table*}[]
\centering
\resizebox{0.9\textwidth}{!}{
\begin{tabular}{l|ccccccc}
\hline
Cycle & Random   & Entropy & Core-Set \cite{DBLP:conf/iclr/SenerS18} & Ensemble \cite{DBLP:conf/cvpr/BeluchGNK18} & MC-dropout \cite{DBLP:conf/icml/GalIG17} & 
PM \cite{DBLP:journals/corr/abs-2103-16130} &
Ours \\ \hline
0   & 25.63$\pm$0.4  & 25.63$\pm$0.4 & 25.63$\pm$0.4  &  27.50$\pm$0.3 &  27.50$\pm$0.3 & \textbf{27.70} $\pm$0.1 & 27.50$\pm$0.3                            \\ 
1   & 28.40$\pm$0.1  & 28.57$\pm$0.2 & 28.10$\pm$0.5  & 28.65$\pm$0.1 &  28.70$\pm$0.2 & 29.28 $\pm$0.1& \textbf{30.07}$\pm$0.4                             \\ 
2   & 29.40$\pm$0.2  & 29.47$\pm$0.1 & 29.57$\pm$0.1 & 29.75$\pm$0.2 &  29.42$\pm$0.2 &30.51$\pm$0.1& \textbf{31.63}$\pm$0.1                             \\ 
3   & 30.20$\pm$0.6  & 30.37$\pm$0.1 & 30.40$\pm$0.4 & 30.43$\pm$0.1 &  30.24$\pm$0.2 &  31.20 $\pm$0.1 &  \textbf{32.10}$\pm$0.1                               \\ 
4   & 31.03$\pm$0.1 & 31.17$\pm$0.2 & 31.17$\pm$0.3 & 31.20$\pm$0.2 & 31.03$\pm$0.1 & 31.86 $\pm$0.1 & \textbf{32.43}$\pm$0.1                           \\ 
5  & 31.47$\pm$0.3 & 31.50$\pm$0.2  & 31.87$\pm$0.2 & 31.75$\pm$0.1 & 31.22$\pm$0.1 & 32.27 $\pm$0.1 & \textbf{32.80}$\pm$0.0                           \\ 
\end{tabular}
}
\caption{\textbf{MS-COCO}. Comparison to state-of-the-art methods. In this case, we initially use $5,000$ randomly sampled images, and, in every active learning cycle, we label $1,000$ extra images. Our method outperforms all the other methods, including ensembles, by a large margin.}
\label{coco-results-a}
\end{table*}

\begin{table*}[!t]
\centering
\begin{tabular}{cc}
\resizebox{0.42\linewidth}{!}{%
\begin{tabular}{l|llll}
\toprule
Cycle & Random                         & SSL-cons. \cite{DBLP:conf/nips/JeongLKK19}                         & SSL-PL \cite{pseudo-labelling}      &  Ours                                                           \\ \hline
0 & 25.63$\pm$0.4 & \textbf{27.50}$\pm$0.3 & \textbf{27.50}$\pm$0.3 & \textbf{27.50}$\pm$0.3\\
1 & 28.40$\pm$0.1 & 28.52 $\pm$0.1 & 28.70 $\pm$0.2& \textbf{30.07}$\pm$0.4\\
2 & 29.40$\pm$0.2 & 29.20 $\pm$0.1& 29.42 $\pm$0.2 &\textbf{31.63}$\pm$0.1\\
3 & 30.20$\pm$0.6 & 30.20 $\pm$0.1& 30.24 $\pm$0.2 & \textbf{32.10}$\pm$0.1\\
4 & 31.03$\pm$0.1 & 31.03 $\pm$0.1& 31.03 $\pm$0.1 & \textbf{32.43}$\pm$0.1\\
5 & 31.47$\pm$0.3 & 31.47 $\pm$0.1& 31.22$\pm$0.1 & \textbf{32.80}$\pm$0.0\\
\end{tabular}
}
&
\resizebox{0.52\linewidth}{!}{%
\begin{tabular}{l|ccccc}
\toprule
Cycle & Random  & Entropy & Inconsistency & Combined  &  Ours \\ \hline
0     & \textbf{27.50}$\pm$0.3 & \textbf{27.50}$\pm$0.3& \textbf{27.50}$\pm$0.3& \textbf{27.50}$\pm$0.3&\textbf{27.50}$\pm$0.3 \\                                       
1    & 28.53$\pm$0.1  & 28.97$\pm$0.3 & 28.90$\pm$0.3& 29.17$\pm$0.2& \textbf{30.07}$\pm$0.4\\  
2     & 29.20$\pm$0.1 & 29.77$\pm$0.1& 29.90$\pm$0.7& 30.17$\pm$0.2& \textbf{31.63} $\pm$0.1\\  
3     & 29.87$\pm$0.1 & 29.93$\pm$0.4 & 30.85$\pm$0.4& 30.90$\pm$0.2& \textbf{32.10} $\pm$0.1\\  
4     & 31.03$\pm$0.1 & 31.10$\pm$0.1& 31.55$\pm$0.1& 31.55$\pm$0.1& \textbf{32.43} $\pm$0.1 \\  
5     & 31.10$\pm$0.1& 31.40$\pm$0.1& 31.65$\pm$0.3& 31.50$\pm$0.1& \textbf{32.80} $\pm$0.0\\  

\end{tabular}
}
\\(a)&(b)
\end{tabular}
\caption{\textbf{MS-COCO}. a) Comparison to two semi-supervised learning methods. We initially use $5,000$ randomly sampled images and, in every other cycle, we label $1,000$ extra images. Our method outperforms both of them by a large margin. b) Ablation study on the effect of entropy, inconsistency, unified score, and our method in MS-COCO. We observe that doing active learning with either entropy or consistency outperforms the semi-supervised model, that the unified score performs better than either of the individual scores, and that our method reaches the best overall results.}
\label{tab:4}
\end{table*}

\section{Detailed results in MS-COCO}
\begin{figure*}[!t]
    \centering
    \begin{tabular}{ccccc}
    \hspace{-0.8cm}\includegraphics[width=0.23\linewidth]{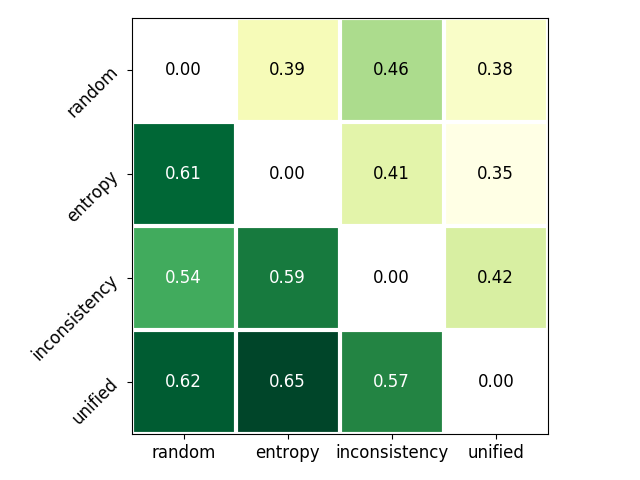}&
    \hspace{-1cm}\includegraphics[width=0.23\linewidth]{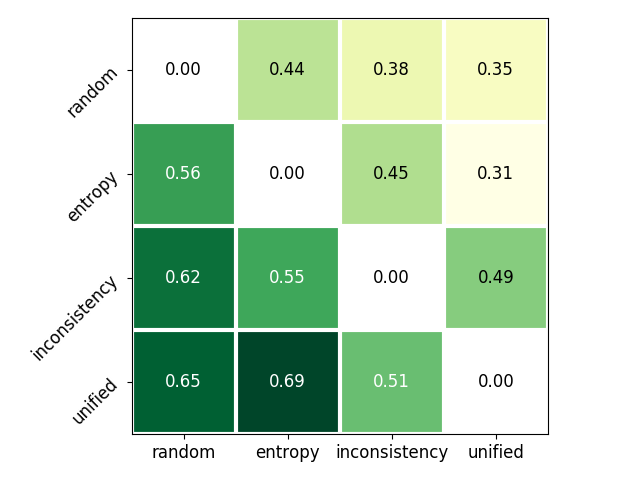}&
    \hspace{-1cm}\includegraphics[width=0.23\linewidth]{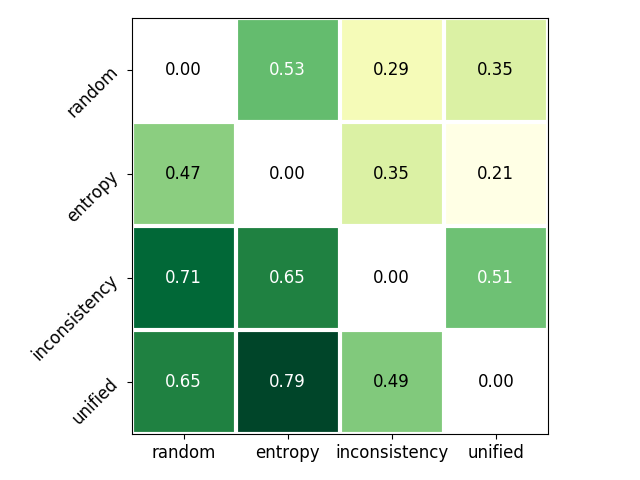}&
    \hspace{-1cm}\includegraphics[width=0.23\linewidth]{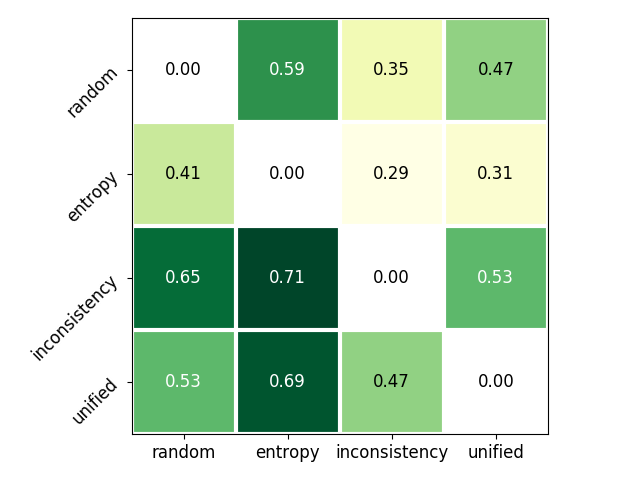}&
    \hspace{-1cm}\includegraphics[width=0.23\linewidth]{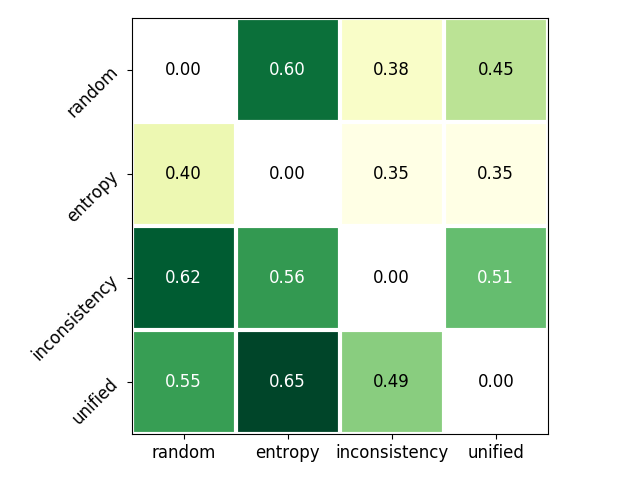}\\
    (1)&(2)&(3)&(4)&(5)
    \end{tabular}
    \caption{\textbf{MS-COCO.} The percentage of classes where one acquisition function outperforms another. Numbers 1-5 represent the active learning cycle. Example: taking the entry "unified" in the y-axis, and "entropy" in the x-axis in (1), we get the value $0.65$ which means that "unified" acquisition function outperforms the "entropy" acquisition function in $65\%$ of classes during the first active learning cycle.}
    \label{acquisition_per_cycle}
    \vspace{1cm}
\end{figure*}

\begin{figure*}[!t]
    \centering
    \begin{tabular}{ccccc}
    \hspace{-0.8cm}\includegraphics[width=0.23\linewidth]{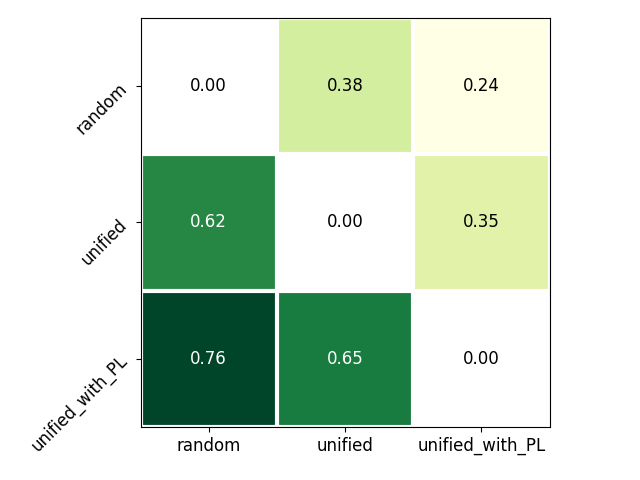}&
    \hspace{-1cm}\includegraphics[width=0.23\linewidth]{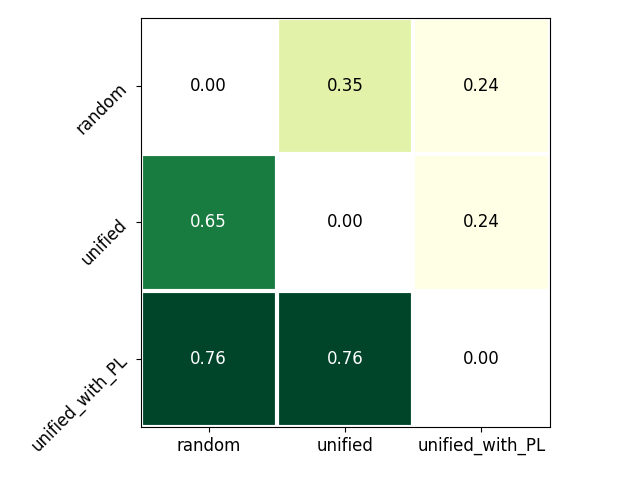}&
    \hspace{-1cm}\includegraphics[width=0.23\linewidth]{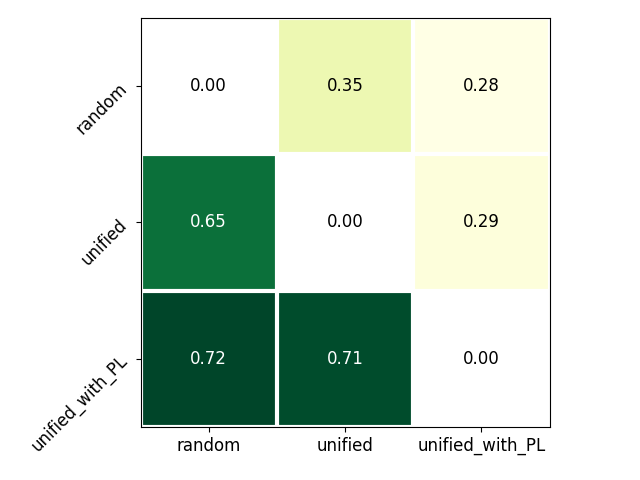}&
    \hspace{-1cm}\includegraphics[width=0.23\linewidth]{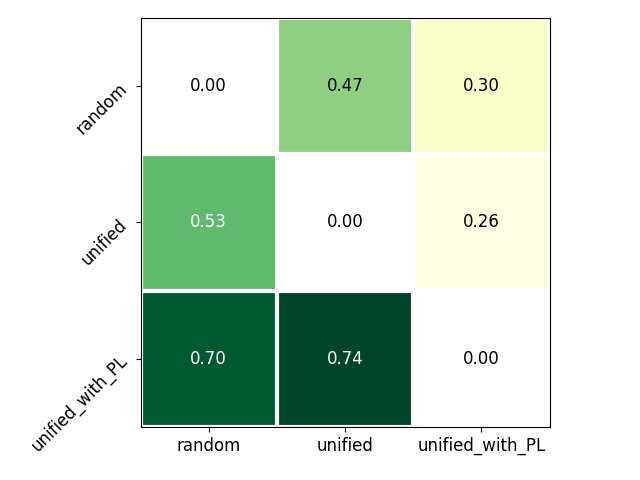}&
    \hspace{-1cm}\includegraphics[width=0.23\linewidth]{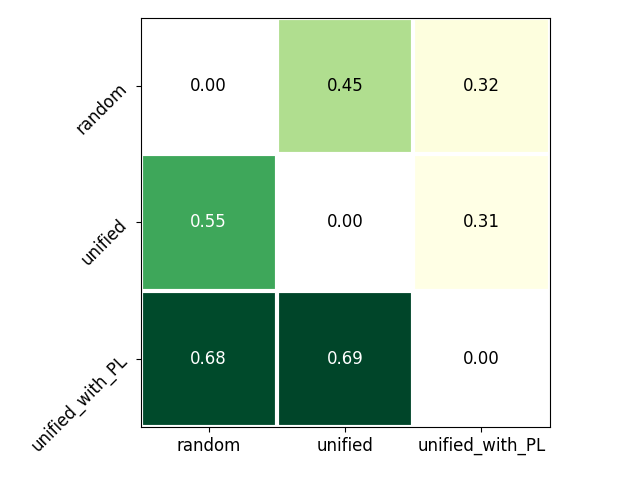}\\
    (1)&(2)&(3)&(4)&(5)
    \end{tabular}
    \caption{\textbf{MS-COCO.} The percentage of classes where our unified acquisition function outperforms random with and without pseudo-labels. Numbers 1-5 represent the active learning cycle. Example: taking the entry "unified\_with\_PL" in the y-axis, and "random" in the x-axis in (1), we get the value $0.76$ which means that our method outperforms the random acquisition function in $76\%$ of classes during the first active learning cycle.}
    \label{acquisition_per_cycle_pl}
\end{figure*}

In the paper we presented aggregated results on the number of classes one acquisition function performs another (lines 688-697), and in pseudo-label performance boost per class (lines 777-783). We presented results by aggregating them over the five cycles of active learning. For completeness, in Figures \ref{acquisition_per_cycle} and \ref{acquisition_per_cycle_pl}, we provide results for each AL cycle in isolation. We see the same trend as in the paper.

\section{Pseudo-labels for class}
In this experiment we analyze different methods for obtaining pseudo-labels. Precisely, instead of obtaining pseudo-labels using the confidence score and a threshold $\tau$ independently of the class, we consider the $k\%$ most confident objects for each class and add them as pseudo-labels. In Table \ref{tab:5}a we present the results where we pseudo-label the top $20\%$, top $30\%$, top $40\%$ most confident predictions for class and compare them with the results of our method described in the main paper. We see that the methods where we pseudo-label per class work well, but worse than our method. Thus, for both simplicity and perormance, we choose to use our class-agnostic method.

\begin{table*}[!t]
\centering
\begin{tabular}{cc}
\resizebox{0.42\linewidth}{!}{%
\begin{tabular}{l|cccc}
\hline
Cycle & Top $20\%$ & Top $30\%$ & Top $40\%$ & Ours  \\ \hline
0  & {63.25}$\pm$0.3 & {63.25}$\pm$0.3 & {63.25}$\pm$0.3 & \textbf{63.25}$\pm$0.3                               \\ 
1  & 69.42$\pm$0.1& 69.63$\pm$0.1& 69.67$\pm$0.3&   \textbf{70.95}$\pm$0.1                             \\ 
2  & 71.84$\pm$0.3& 71.84$\pm$0.3& 72.14$\pm$0.2&  \textbf{72.88}$\pm$0.1                             \\ 
3  & 73.34$\pm$0.3& 73.47$\pm$0.1& 73.56$\pm$0.2& \textbf{73.55}$\pm$0.1                               \\ 
4  & 74.53$\pm$0.1& 74.53$\pm$0.2& 74.32$\pm$0.2& \textbf{74.75}$\pm$0.1                           \\ 
5  & 75.13$\pm$0.1& 75.39$\pm$0.2& 75.19$\pm$0.2&  \textbf{75.60}$\pm$0.1                           \\ 
\end{tabular}
}
&
\resizebox{0.37\linewidth}{!}{%
\begin{tabular}{l|cccc}
\hline
Cycle & Random & Balanced quarter & Unified    \\ \hline
0  & 53.66$\pm$0.2 & \textbf{63.37}$\pm$0.2 & 63.25$\pm$0.2                         \\ 
1  & 67.39$\pm$0.4 & 68.12$\pm$0.2 & \textbf{68.40}$\pm$0.3                              \\ 
2  & 69.90$\pm$0.5 & 70.12$\pm$0.6 & \textbf{70.84}$\pm$0.8                              \\ 
3  & 71.38$\pm$1.2 & 71.92$\pm$0.3 & \textbf{72.93}$\pm$0.3                             \\ 
4  & 73.53$\pm$0.4 & 73.48$\pm$0.3 & \textbf{73.66}$\pm$0.2                             \\ 
5  & 74.30$\pm$0.4 & 73.90$\pm$0.4 & \textbf{74.98}$\pm$0.5                           \\ 
\end{tabular}
}
\\(a)&(b)
\end{tabular}
\caption{\textbf{VOC07+12}. a) The results of adding top $k\%$ most confident pseudo-labels for class, compared to the results of our method. \textit{Top $20\%$}, \textit{Top $30\%$}, \textit{Top $40\%$} represent the methods where we choose to pseudo-label the most confident $20\%$, $30\%$ and $40\%$ pseudo-labels per class. \textit{Ours} represent our method where we pseudo-label all the objects for which the network's confidence is greater than $0.99$. b) Accuracy as a function of label/unlabeled sampling strategy. \textit{Random} refers to random sampling from the entire dataset, \textit{Balanced quarter} refers to having a quarter of labeled samples; \textit{Unified} refers to half of the samples being labeled. Our balanced strategy outperforms the other two strategies. Note that in order to check only the effect of balancing, we do not add pseudo-labels during the training.} 
\label{tab:5}
\end{table*}

\section{Engineering tricks to consider}

\subsection{Non-maximum suppression}

We found the effect of non-maximum suppression (NMS) to be very important in all AL methods. Without applying NMS, active learning methods did not work better than a random sampling method. We hypothesize that this happens because if we do not apply NMS, the number of detected boxes is in the hundreds, so by sheer chance, some of them might have high acquisition scores. Considering that in a real-world scenario these boxes would be \textit{killed} by NMS, we conclude that these boxes should not be used to compute an acquisition score. Thus, for every image, we apply NMS before proceeding with the computation of the acquisition score.

\subsection{Balanced mini-batches}

In the main paper, for every experiment, we force that half of the samples in a mini-batch are labeled. In this experiment, we evaluate the effect of varying the number of labeled samples in a mini-batch. In particular, we compare our results to having only half of the samples labeled, and a random approach. In order to be able to quantify the effect of balancing, we do all the experiments without adding pseudo-labels. We present the results in Table \ref{tab:5}b. We observe that our strategy of balancing the mini-batches so they contain an equal number of labeled and unlabeled samples, performs best by up to $1pp$ in all AL cycles except the zeroth one, when it gets outperformed by $0.12pp$ by the strategy where only a quarter of samples contain labels. We also observe that the strategy where we do only random sampling consistently reaches the worst results. In fact, in the zeroth AL cycle it gets outperformed by the balanced strategies by almost $10pp$. This can be explained by the fact that the number of labeled samples ($2,000$) is much lower than the number of unlabeled samples ($14,651$), so in a mini-batch of size $32$, in average, only $3.86$ samples have labels. In some mini-batches, the number of labeled samples is $0$, and thus the loss function becomes completely self-supervised. We observe that when the number of labeled samples increases (by labeling other images during AL stage), the overall performance increases, but it still lags behind the the balanced strategies.

\end{document}